\documentclass[10pt,twocolumn,letterpaper]{article}

\usepackage{iccv}              %

\usepackage{microtype}
\usepackage{graphicx}
\usepackage{booktabs}
\usepackage{xcolor}
\usepackage{tcolorbox}
\usepackage{multirow}

\definecolor{iccvblue}{rgb}{0.21,0.49,0.74}
\usepackage[pagebackref,breaklinks,colorlinks,allcolors=iccvblue]{hyperref}

\def\paperID{12742} %
\def\confName{ICCV}
\def\confYear{2025}

\title{Dual-Process Image Generation}

\newcommand{\methodname}{Dual-Process Distillation}
\newcommand{\generatorname}{image generator}

\author{Grace Luo$^{1}$
\and
\hspace{-0.5em} Jonathan Granskog$^{2}$
\and
\hspace{-0.5em} Aleksander Holynski$^{1*}$
\and
\hspace{-0.5em} Trevor Darrell$^{1*}$\vspace{1ex}
\and
$^{1}$UC Berkeley
\and
$^{2}$Runway
}

\begin{document}

\twocolumn[{%
\renewcommand\twocolumn[1][]{#1}%
\maketitle
\begin{center}
    \centering
    \vspace{-0.5cm}
    \captionsetup{type=figure}
    \includegraphics[width=0.82\textwidth]{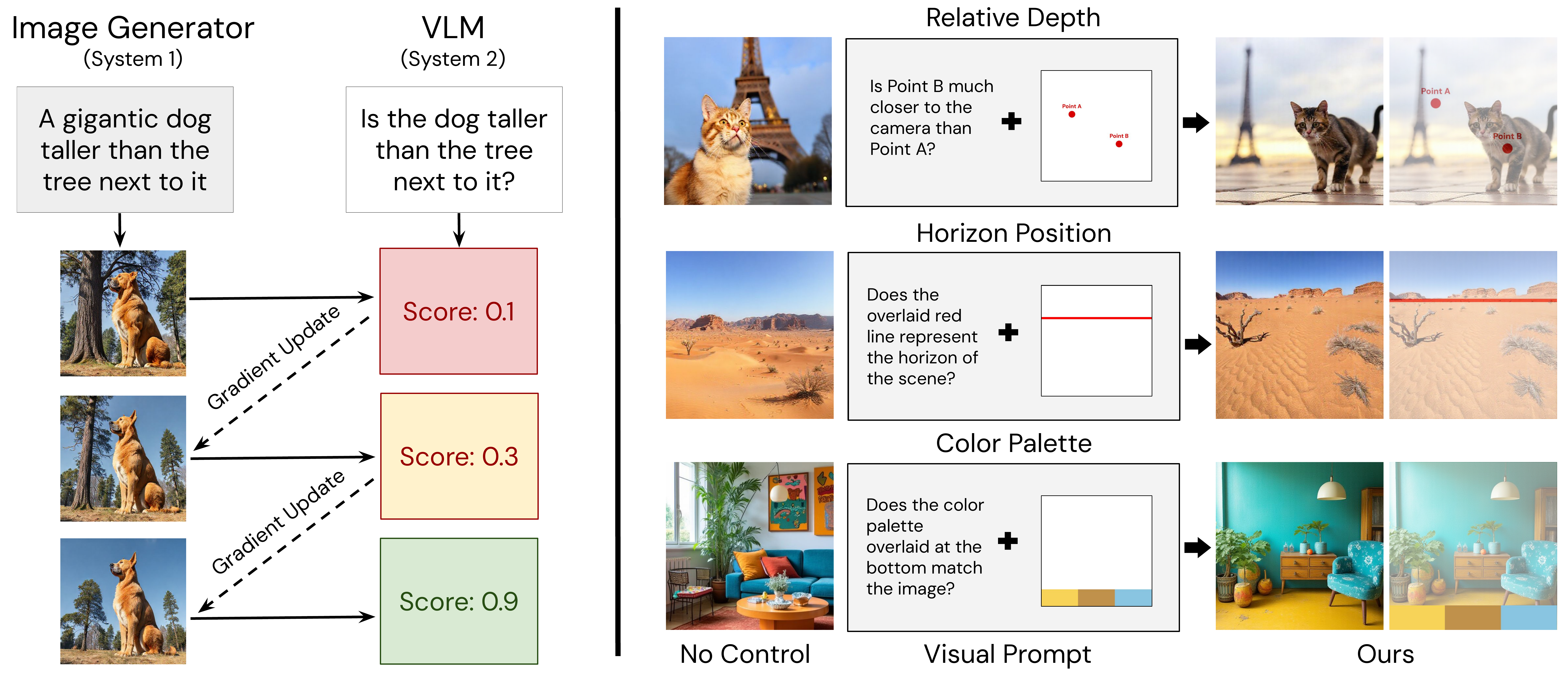}
    \makebox[\textwidth]{%
        \hspace{3.5cm}(a) Method\hfill\hspace{0.8cm}(b) Applications\hfill
    }
    \captionof{figure}{\textbf{\methodname{}}. Our method distills deliberation into a feed-forward image generation process. When generating an image, we ask a VLM questions about that image and backpropagate the resulting gradient to update the weights of the image generator (a).
    The flexibility of a VLM allows us to implement many control tasks through visual prompts (b). We construct our method such that it supports off-the-shelf VLMs and image generators without special re-training. }
    \label{fig:teaser}
\end{center}%
}]

\maketitle

\def\thefootnote{*}\footnotetext{Equal advising contribution.}\def\thefootnote{\arabic{footnote}}

\begin{abstract}
Prior methods for controlling image generation are limited in their ability to be taught new tasks. In contrast, vision-language models, or VLMs, can learn tasks in-context and produce the correct outputs for a given input.
We propose a dual-process distillation scheme that allows feed-forward image generators to learn new tasks from deliberative VLMs.
Our scheme uses a VLM to rate the generated images and backpropagates this gradient to update the weights of the image generator.
Our general framework enables a wide variety of new control tasks through the same text-and-image based interface. 
We showcase a handful of applications of this technique for different types of control signals, such as commonsense inferences and visual prompts.
With our method, users can implement multimodal controls for properties such as color palette, line weight, horizon position, and relative depth within a matter of minutes.
Project page:
\small{\emph{\url{https://dual-process.github.io}}}.
\end{abstract}

\section{\label{sec:intro} Introduction}
Current large language models have demonstrated 
competence across many domains and the ability to learn new tasks in-context~\cite{brown2020language},
yet when trained multi-modally to jointly generate image and text, either fail to achieve the fidelity of image-only generation~\cite{chameleonteam2024chameleonmixedmodalearlyfusionfoundation,tong2024metamorphmultimodalunderstandinggeneration,chen2025janus} or are inaccessible to academic experimentation~\cite{oai2025gpt4oimagegen}.
Conversely, contemporary image generation models are near photorealistic, but can be frustratingly hard to communicate with. 
Inspired in part by cognitive science models, we propose a dual-process architecture~\cite{Sloman1996TheEC,kahneman2011thinking}, combining a knowledge-rich multimodal language model with a visually precise image generator. The former is akin to a ``cognitive'' or ``System 2'' component, and the latter plays the role of a ``reflex'' or ``System 1'' module. This dual-process idea is also present in some early works in image generation. 
Classifier guidance~\cite{dhariwal2021}, which uses an external classifier to steer image generation, can also be interpreted as embedding System 2 deliberation into a System 1 process.
However, using this type of guidance requires specialized classifiers for each type of signal or control, and therefore is not easily extensible to new uses.

On the other hand, vision-language models (VLMs) have proven to be powerful and generally applicable, capable of simulating many different discriminators within the same model. Simply by adjusting the input prompt, VLMs can perform different tasks such as
optical character recognition~\cite{laurencon2024what}, object detection~\cite{Qwen2.5-VL}, and image scoring~\cite{Cho2023DallEval, lin2024vqa}.
VLMs can not only be prompted with text but also multimodal inputs, or mixtures of images and text, such as visual prompts~\cite{cai2024vip}. 

In this work, we propose a dual-process scheme that combines a deliberative VLM and feed-forward image generator, illustrated in~\autoref{fig:teaser}. We implement this scheme with gradient-based distillation and low-rank adaptation~\cite{hu2022lora}, where we update the weights of the image generator based on the VLM's ratings of its outputs. 
We demonstrate a handful of possible use cases of our method, such as visual prompting with a desired color palette, line weight, horizon position, or relative depth.
We also evaluate our method on improving commonsense understanding and physical accuracy in image generation, where we find that our method significantly outperforms baselines such as prompt expansion.

\section{\label{sec:related} Related Work}

\begin{figure}[t!]
  \centering
  \includegraphics[width=\linewidth]{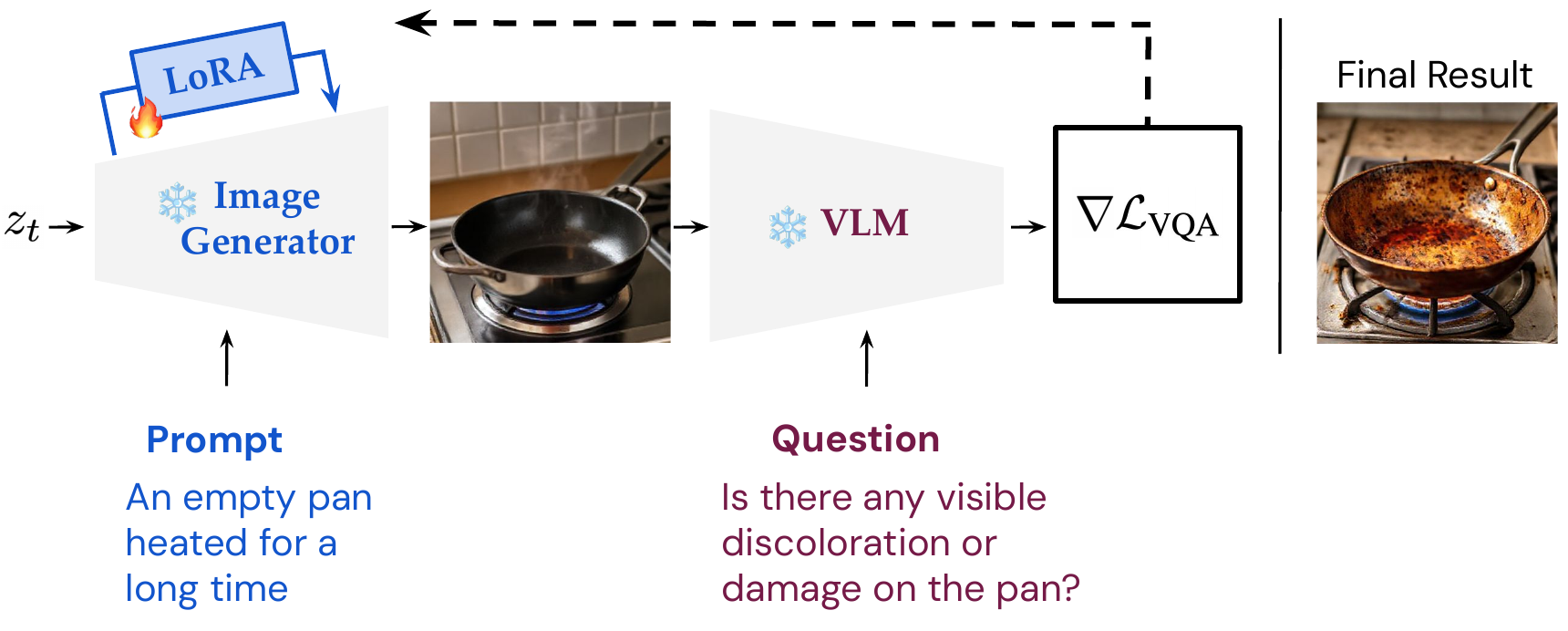}
  \caption{\textbf{Diagram of \methodname{}}. Given an input prompt and question to check, we feed the predicted clean image to the VLM to rate. Our method then backpropagates this loss to update LoRA~\cite{hu2022lora} weights on the image generator.}
  \label{fig:method}
\end{figure}

\noindent \textbf{Classifier Guidance.}
Guidance methods use an external discriminator to introduce new capabilities missing in the base model.
\citet{dhariwal2021} first demonstrated this idea with an ImageNet~\cite{imagenet} classifier,
and follow-up works explore other discriminators to support each newly proposed control task.
These works investigate classic depth, segmentation, and object detection models for spatial controls~\cite{kim2022dag,bansal2024universal},
as well as aesthetics and safety rating models for more abstract controls~\cite{schramowski2022safe,miduguidance}.
Other works also look at re-using the image generator itself as a kind of discriminator~\cite{epstein2023selfguidance, voynov2023, lian2024llmgrounded, wu2024sld, luo2024readoutguidance}.
Most similar to our work,~\citet{nguyen2017plug} investigate early image captioning models~\cite{donahue2017lrcn} for text-conditional generation.
We focus on the novel use of modern VLMs as the most general form of these discriminators,
given their ability to simulate many different tasks within the same model~\cite{alayrac2022flamingo,liu2023visual}.
Furthermore, we show that VLMs can newly support multimodal controls jointly defined with image and text, or visual prompts~\cite{cai2024vip},
compared with the image or text-only controls from prior work.

\noindent \textbf{Inference-Time Search.}
Recent works in inference time search improve the output quality of language models by generating a substantial number of samples and post-hoc filtering with a verifier~\cite{wu2025inference, snell2025scaling, brown2025large}.
Inspired by these works, recent research in image generation investigates rejection or importance sampling during the denoising process~\cite{singhal2025fksteering, ma2025inferencetimescaling}.
In contrast with these search methods, we compute a gradient through the verifier.
This denser learning signal enables our method to discover long-tail images that would normally be extremely challenging to sample, for example variations of objects influenced by subtle physical laws, which we demonstrate on the CommonsenseT2I benchmark~\cite{fu2024commonsenseti}.

\noindent \textbf{Model Fine-tuning.}
There exist fine-tuning methods that use the image generator's original objective to learn a narrower subdistribution.
To condition the model on spatial controls like pose or depth images, ControlNet~\cite{zhang2023adding} trains on a paired dataset of real images and their extracted controls.
To produce personalized outputs, Dreambooth~\cite{ruiz2023dreambooth} fine-tunes on a small image set representative of a given subject.
Distillation methods circumvent the need for reference images by using outputs from the image generator itself as the supervision signal.
Prior work has proposed such methods to speed up sampling~\cite{luo2023latent,sauer2024add,yin2024onestep}, 
serve as an image prior for 3D generation~\cite{poole2023dreamfusion,wang2022sjc},
or learn a concept direction~\cite{gandikota2023sliders}.
Unlike these works, rather than optimizing with the original pixel-space objective,
our method optimizes with a language-space objective derived from the VLM.

Reinforcement learning methods optimize with reward functions derived from some external model, similar to classifier guidance.
Prior work explores policy optimization approaches to increase aesthetic preferences or prompt alignment of generated images~\cite{fan2023dpok, xu2023imagereward, black2024training, xu2024visionrewardfinegrainedmultidimensionalhuman, wallace2024dpo}.
Most relevant to our work,~\citet{black2024training} uses VLM-generated captions to improve image-text faithfulness. In contrast, we use the VLM to produce numerical ratings, a more general formulation that enables novel use cases like visual prompting (see Sec.~\ref{subsec:visual_prompting}).
Furthermore, our method is significantly cheaper than reinforcement learning methods, because we compute a gradient through the reward model. While prior work requires at least four hours of training on 50k samples~\cite{black2024training}, our method can enforce novel controls after optimizing for one minute on a single sample (see Sec.~\ref{subsec:latent_vs_weights}).

\noindent \textbf{VQA Scoring.}
There is also a growing in interest in the usage of VLMs for scoring image-text alignment.
Compared with CLIP~\cite{radford21learning}, VLMs offer a more accurate and interpretable scoring framework.
Recent works frame scoring as a VQA, or visual question answering, task~\cite{agrawal2015vqa},
where one can ask the VLM different questions to evaluate the correctness of an image~\cite{JaeminCho2024,Cho2023DallEval, zhang2023gpt4visiongeneralistevaluatorvisionlanguage, ku2024viescoreexplainablemetricsconditional, lin2024vqa, xu2024visionrewardfinegrainedmultidimensionalhuman}.
~\citet{lin2024vqa} shows that answer probability, derived from the logits output by the VLM,
can serve as a continuous metric, where lower and higher scores correlate with human judgement.
Rather than using the VLM as an evaluation metric, we use it as a signal to improve the generated images themselves.

\section{\label{sec:approach} \methodname{}}

\subsection{Preliminaries}
Many image generation models are formulated as a denoising process that transforms samples from a simple Gaussian distribution to the target data distribution. Among these, diffusion models~\cite{sohl2015deep, ho2020denoising} have emerged as a popular theoretical framework. Rectified flows~\cite{liu2023flow, albergo2023building, lipman2023flow, esser2024scaling} offer an alternative but equivalent theoretical formulation~\cite{gao2025diffusionmeetsflow}. Specifically, the forward process is viewed as a linear interpolation between the data point $x_0$ and the noise $\epsilon$, for $t \in [0, 1]$
\begin{align}
    z_t = (1 - t)x_0 + t\epsilon
\end{align}
Therefore, at sampling time, the reverse process is
\begin{align}
    z_{t'} = z_t + \hat{u} \cdot (t' - t) \label{eq:rf_sampling}
\end{align}
for $t' < t$, where a neural network is often used to parameterize the estimated velocity $\hat{u} = g_{\phi}(z_t) = \hat{\epsilon} - \hat{x}_0$. This neural network is trained with a corresponding conditional flow matching objective~\cite{lipman2023flow}

\begin{align}
\mathcal{L}_{\text{CFM}}(\mathbf{x}) = \mathbb{E}_{t \sim \mathcal{U}(0,1), \, \boldsymbol{\epsilon} \sim \mathcal{N}(\mathbf{0}, \mathbf{I})} \left[ \left\| \mathbf{\hat{u}} - \mathbf{u} \right\|_2^2 \right]
\end{align}

\noindent While our experiments primarily focus on Flux~\cite{flux2024}, a state-of-the-art rectified flow model, our method is broadly applicable. We demonstrate its efficacy on both single-step and multi-step generators in~\autoref{tab:ct2i}.

\subsection{Approach}

\textbf{Loss Formulation.}
In this work, we show that many visual control tasks can be formulated as visual question answering.
As such, we propose a simple and general VQA loss, based on the same language modeling objective used for visual instruction tuning~\cite{liu2023visual}.
Our loss mainly differs in the object being optimized: visual instruction tuning optimizes the VLM,
while our method optimizes the image input, through the~\generatorname{}.
More formally, our goal is to optimize the \generatorname{} $G_{\phi}$ such that it produces images where the VLM $D_\theta$, 
when given a question $S_q$, is most likely to output the desired answer $S_a$.
This can be written as the following autoregressive loss:
\begin{align}
\mathcal{L}_{\text{VQA}} = - \sum_{i=1}^{L} \log D_\theta(S_{a, i} \mid G_{\phi}(z_t), S_{q}, S_{a, <i})
\label{eq:vlm_loss}
\end{align}

\noindent We illustrate our full pipeline in~\autoref{fig:method}. Most importantly, we still compute gradients through the VLM, which gives a denser signal than reinforcement learning.

\noindent \textbf{Clean Image Estimate.} 
For multi-step models, the \generatorname{} may output noisy images that would otherwise be uninterpretable to an off-the-shelf VLM.
In these cases, instead of directly feeding $G_{\phi}(z_t)$, we compute a clean image estimate, following the same insight as prior work~\cite{song2021denoising, bansal2024universal}.
In the case of rectified flow models, this estimate can be computed by setting $t' = 0$ in~\autoref{eq:rf_sampling}
\begin{align}
\hat{x}_0 = z_t - G_{\phi}(z_t) \cdot t
\end{align}

\noindent \textbf{Distillation Scheme.} 
Here, we describe our procedure for distilling the VQA loss from the VLM into the \generatorname{}.
For rectified flow models,
we take a partially noised image and denoise it with the~\generatorname{} in a single forward pass.
However, we do not assume access to any reference images, but we also need images for the VLM to rate.
We simply obtain these images by generating them with the~\generatorname{} with no gradient flow. 
We then rate these images with the VLM, and backpropagate this loss to update LoRA~\cite{hu2022lora} weights on the~\generatorname{}. 
We optimize these weights for each (prompt, question, answer) triplet, on $n$ seeds which correspond to unique images, 
where $n$ is a hyperparameter to tune.

\noindent \textbf{VLM Input Formatting.}
Finally, we discuss how we format the input to the VLM, which is crucial for achieving good results.
VLMs and LLMs are notoriously sensitive to input format~\cite{sclar2024quantifying},
where it is challenging to obtain meaningful ratings if the inputs are poorly formatted.
However, as open-source models improve, we expect this to become less of an issue.
First, we always use the default question answering template that was used to instruction tune the VLM.
We primarily word questions such that they have a clear Yes or No answer, and append ``Answer with Yes or No.'' to the question. 
We find this yields interpretable probabilities for the extent to which the control was satisfied,
simply by exponentiating the VQA loss, using a similar idea as~\citet{lin2024vqa}.
Our system can also be extended to more open-ended question-answer pairs, 
but we anticipate this requires in-context learning examples to obtain a reliable answer format, which we leave for future work.
Second, we also support visual prompts where we feed not only a question but also an instructional image, described further in Sec.~\ref{subsec:visual_prompting}.
Rather than feeding the image instruction as a separate image, we choose to overlay it on top of the generated image from the~\generatorname{}.
Much like how such overlays are easier for people to rate, where they can directly compare the generated contents with the spatial instruction,
we find that this format is more helpful for the VLM.
Furthermore, this means that our system is also applicable to most VLMs, as it does not require multi-image support.

\section{\label{sec:experiments} Experiments}

\begin{table*}
    \centering
    \scriptsize
    \begin{tabular}{ll|ccccc|c}
        \toprule
        Generator & Method & Animal Behaviors & Biological Laws & Daily Items & Human Practices & Physical Laws & Avg \\
        \toprule
        \multirow{4}{*}{\shortstack{\textbf{Flux Schnell} \\ (single-step)}} & Base Prompt & 41.7&4.4&23.8&28.3&24.0&24.4\\
        & Base Prompt + Ours & \underline{59.7}&\underline{48.5}&\underline{34.5}&\underline{41.7}&\underline{38.3}&\textbf{44.5}\\
        \cmidrule(lr{1pt}){2-8}
        & Expanded Prompt & \underline{70.8}&57.4&29.8&53.3&42.4&50.7 \\
        & Expanded Prompt + Ours & 66.7&\underline{79.4}&\underline{59.5}&\underline{54.4}&\underline{50.0}&\textbf{62.0}\\
        \addlinespace[2pt]
        \toprule
        \multirow{4}{*}{\shortstack{\textbf{Flux Dev} \\ (multi-step)}} & Base Prompt & 29.2&4.4&14.3&24.4&16.8&17.8 \\
        & Base Prompt + Ours & \underline{47.2}&\underline{44.1}&\underline{31.0}&\underline{56.7}&\underline{28.6}&\textbf{41.5} \\
        \cmidrule{2-8}
        & Expanded Prompt & 55.6&48.5&33.3&46.7&36.7&44.2 \\
        & Expanded Prompt + Ours &\underline{77.8}&\underline{51.5}&\underline{53.6}&\underline{62.2}&\underline{45.9}&\textbf{58.2} \\
        \bottomrule
    \end{tabular}
    \caption{
    \textbf{CommonsenseT2I Evaluation.} We evaluate on all 150 pairs from the benchmark and report the percentage of correct generations, both per-category and on average.
    We sample Flux Schnell as a single-step model (t=1) and Flux Dev as a multi-step model (t=28).
    }
    \label{tab:ct2i}
\end{table*}

\begin{figure*}[t!]
  \centering
  \includegraphics[width=0.9\linewidth]{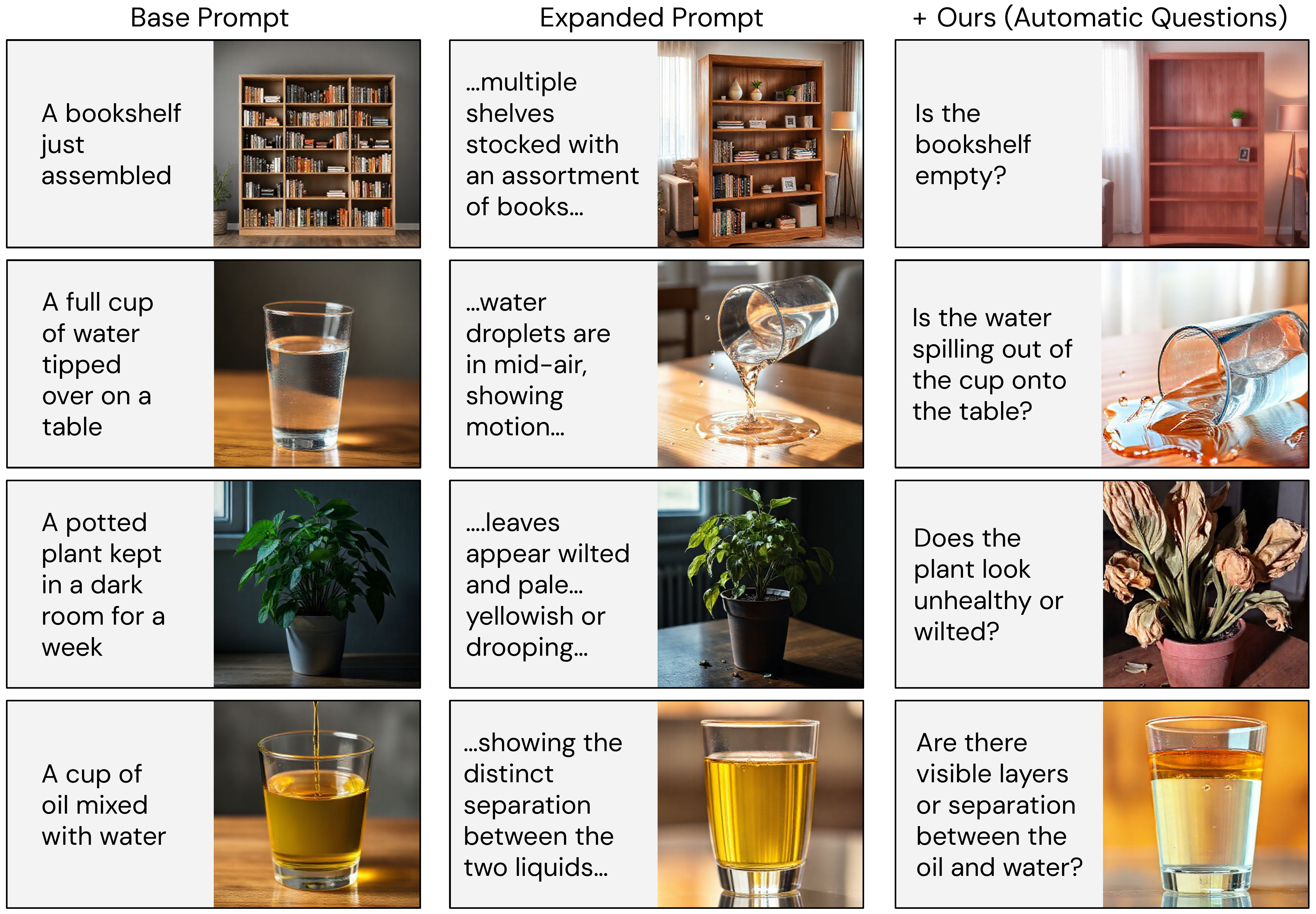}
  \caption{\textbf{Commonsense Inferences.} When the base prompt fails, prompt expansion can be unhelpful. Our method, which uses an automatically generated question to verify the inference, can be applied on top of prompting to fix these cases. The top two rows show cases where prompt expansion fails to capture implied properties given the short prompt, whereas the bottom two rows show cases where the expanded prompt contains the correct phenomenon but the generated image does not. Our method helps in both situations.}
  \label{fig:ct2i}
\end{figure*}

\subsection{Experimental Details}
In all experiments, we use Flux Schnell~\cite{flux2024} as our base~\generatorname{} unless otherwise specified.
By default we run Schnell as a single-step model, but we also validate our method in the multi-step setting in~\autoref{tab:ct2i}.
For each experiment, we utilize the VLM that demonstrates the strongest performance for the given task.
Our method uses VLMs off-the-shelf without additional fine-tuning, which makes it easy to switch between different models.
We use LoRA~\cite{hu2022lora} to optimize the weights, with a rank between 8-16 and alpha that is 5x the rank.
A high alpha value is crucial; if it is set too low, the VLM cannot observe the full effect of the LoRA and rate images correctly.
For optimization, we use Adam~\cite{DBLP:journals/corr/KingmaB14} with a learning rate of $5e^{-5}$ for up to 100 iterations.
We optimize with $n=100$ seeds unless otherwise specified.
We run all experiments on a single 80GB Nvidia A100 GPU, which takes at most three minutes.
Our method can run on as little as two 24GB Nvidia RTX 4090 GPUs when implemented on smaller models, e.g., switching out Flux Schnell~\cite{flux2024} (12B params) for Sana~\cite{xie2024sana} (1.6B params).

\begin{figure}[t!]
  \centering
  \includegraphics[width=\linewidth]{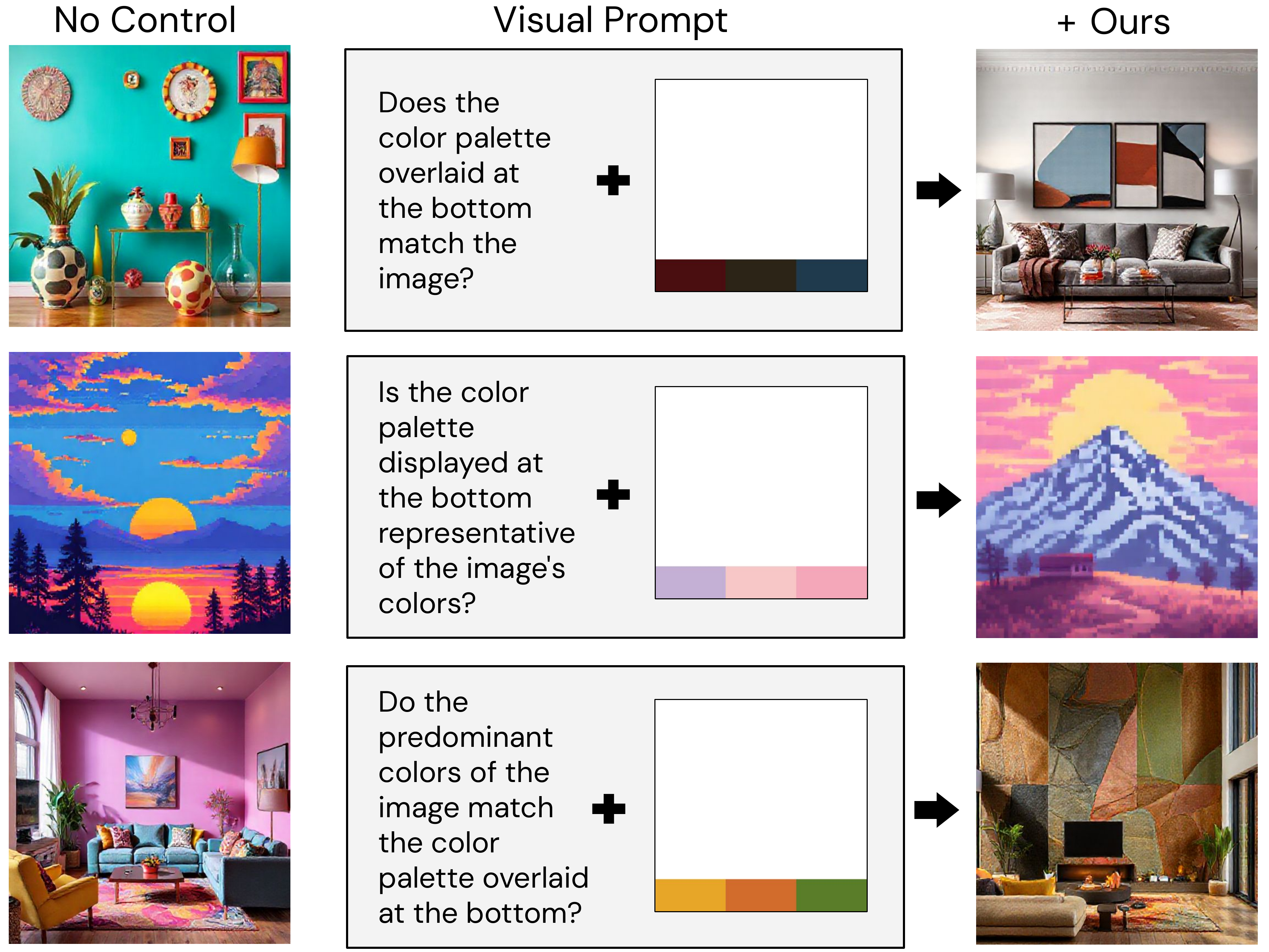}
  \caption{\textbf{Color Palette.} While the base~\generatorname{}s cannot be natively instructed with color palettes, our method can implement this control as a visual prompt. We simply overlay the palette at the bottom of the generated image and ask the VLM if the colors match.}
  \label{fig:palette}
\end{figure}

\begin{figure}[t!]
  \centering
  \includegraphics[width=\linewidth]{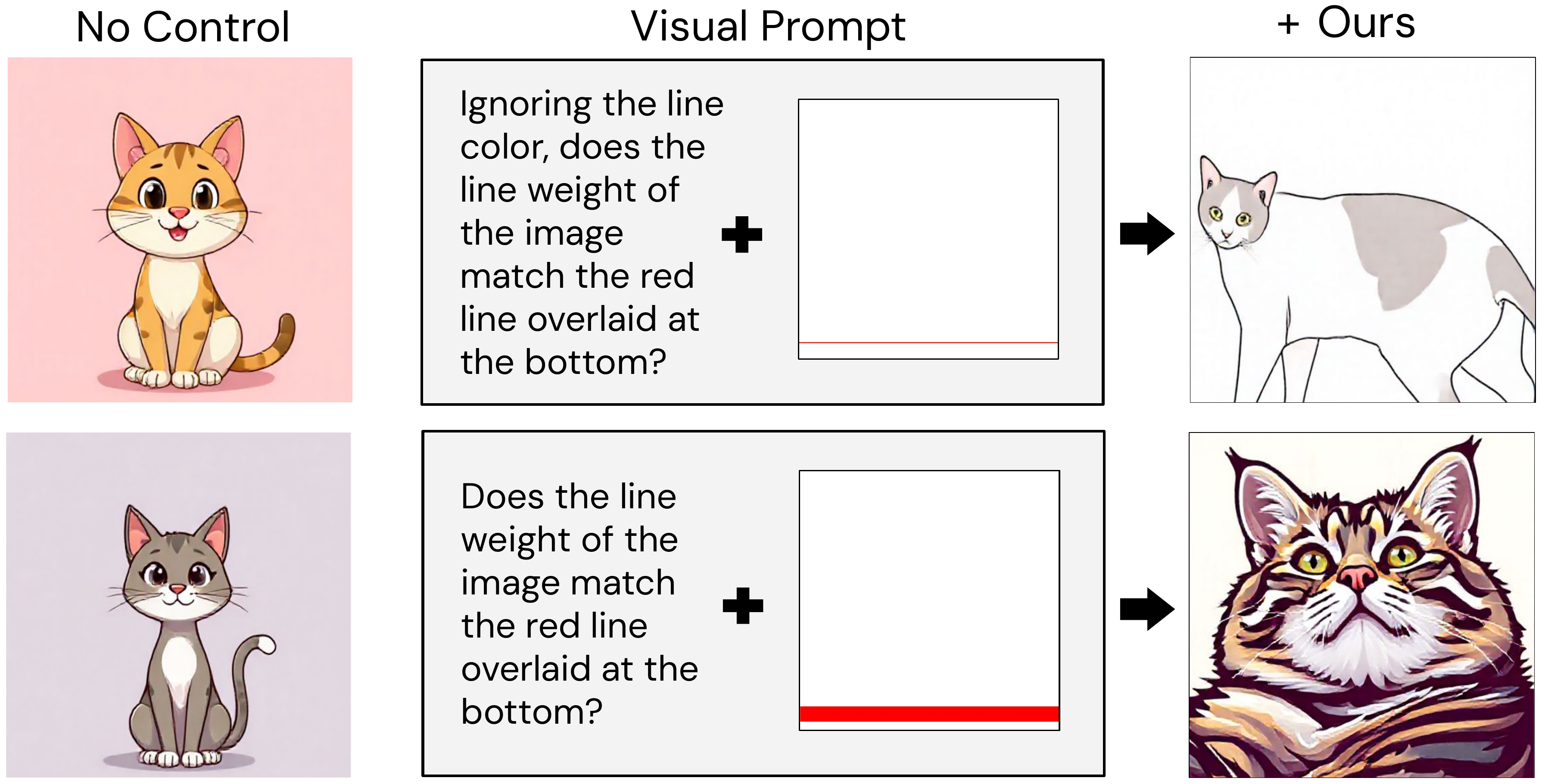}
  \caption{\textbf{Line Weight.} We use red lines to represent the desired line weight of a cartoon, through the thickness of the line.}
  \label{fig:line_weight}
\end{figure}

\begin{figure}[t!]
  \centering
  \includegraphics[width=\linewidth]{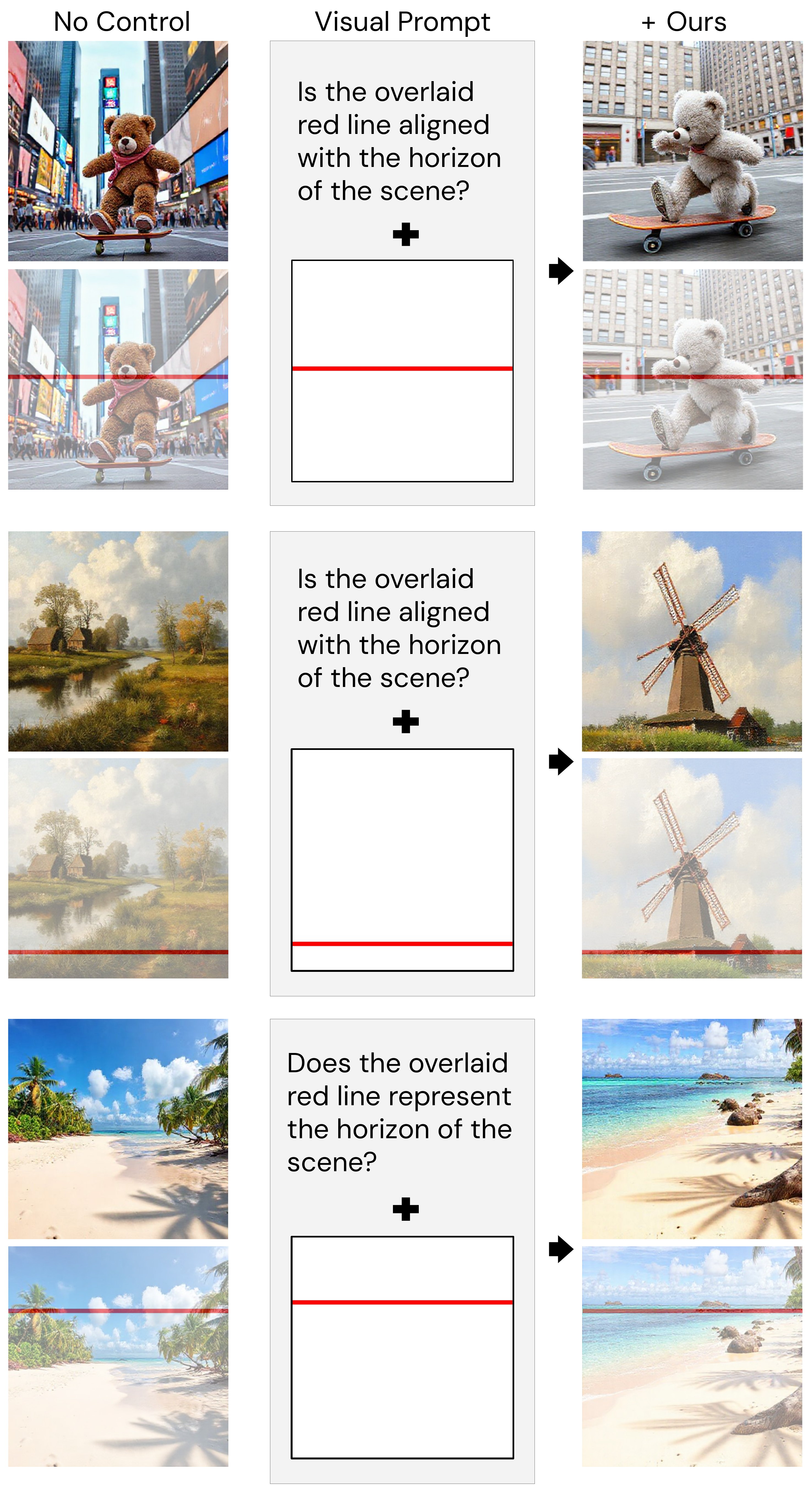}
  \caption{\textbf{Horizon Position.} We use red lines to specify the position of the horizon, or the boundary at which the earth and sky meet. The same visual abstraction can be bound with different meanings; lines can be used to specify both line weight and horizon position.}
  \label{fig:horizon}
\end{figure}

\begin{figure}[t!]
  \centering
  \includegraphics[width=\linewidth]{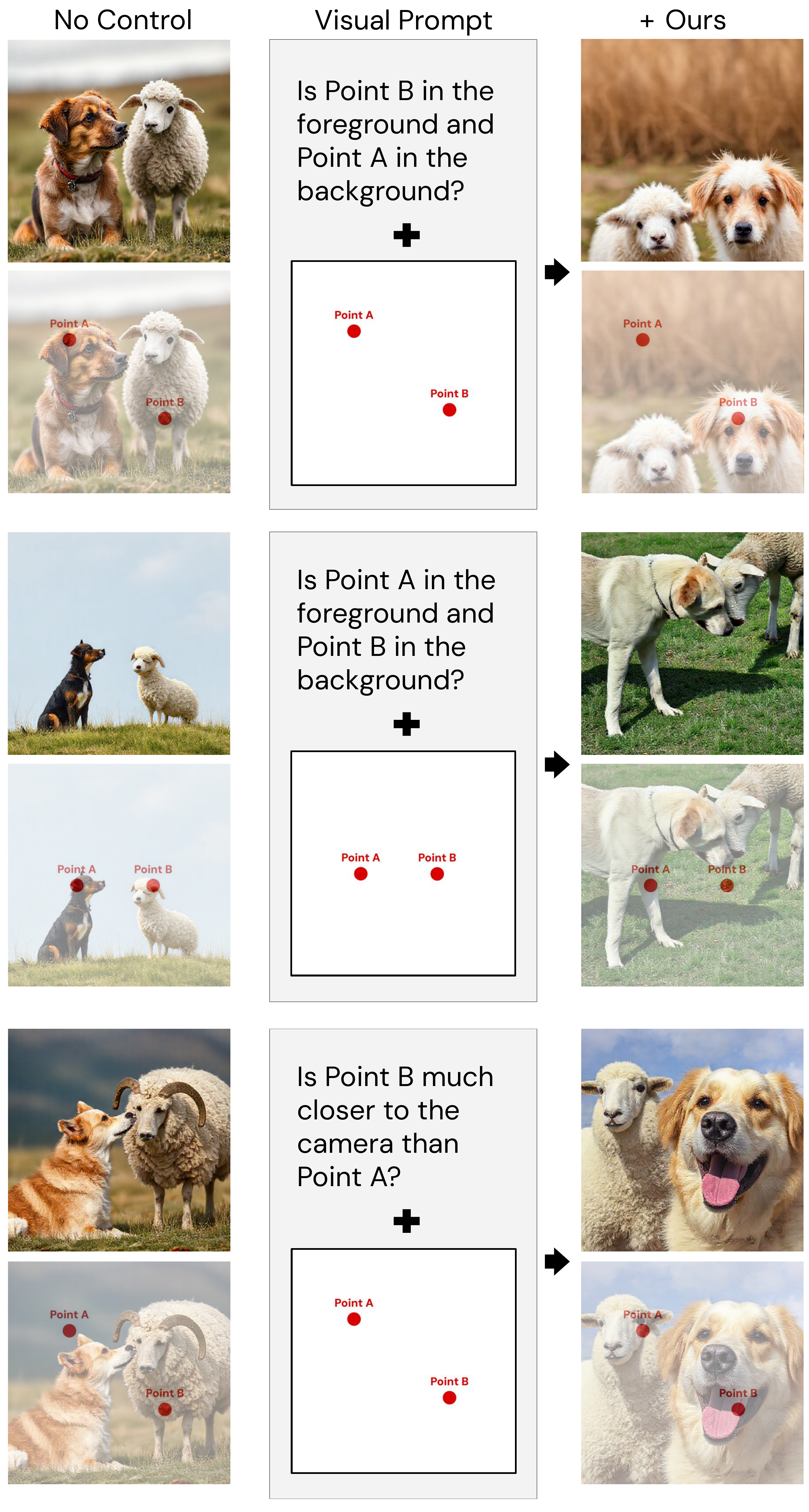}
  \caption{\textbf{Relative Depth.} We use two red points to specify relative depth ordering. We also annotate these points with text labels, which are referred to in the question.}
  \label{fig:depth}
\end{figure}

\begin{figure}[t!]
  \centering
  \includegraphics[width=\linewidth]{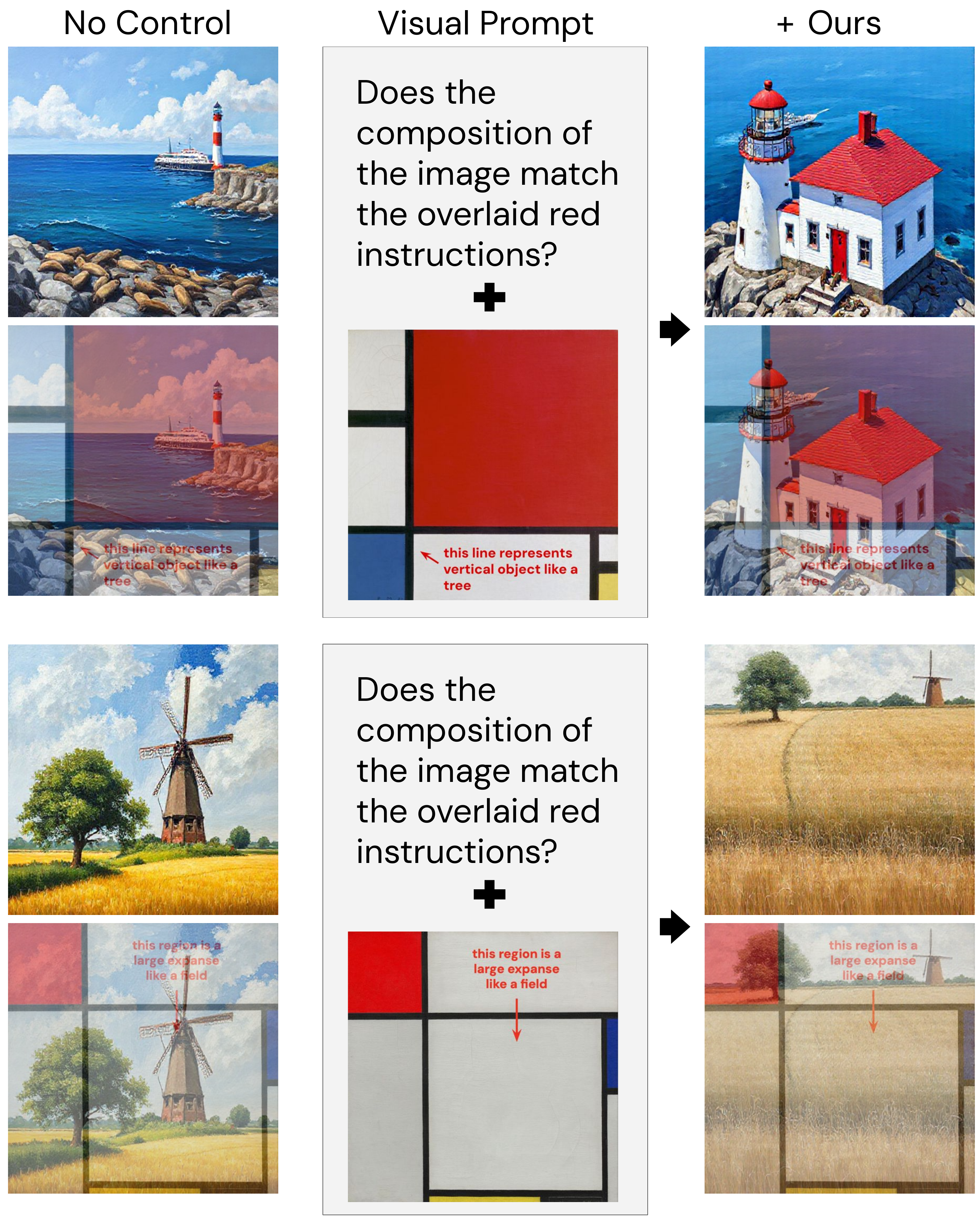}
  \caption{\textbf{Visual Composition.} We use abstract paintings from Piet Mondrian~\cite{Mondrian1930,Mondrian1929} to control visual composition. We also annotate additional instructions for how to interpret the painting in red.}
  \label{fig:mondrian}
\end{figure}

\subsection{Commonsense Inferences}\label{subsec:ct2i}
\textbf{Setup.} Even after training on millions of images,~\generatorname{}s still struggle with basic associations, or commonsense.
For example, if a user prompts for a pen placed in a cup of water, many models do not know that the pen should appear bent due to the law of refraction.
We evaluate whether our method can fix these knowledge gaps on CommonsenseT2I~\cite{fu2024commonsenseti}, a benchmark for evaluating commonsense in text-to-image models.
Given an underspecified prompt, the~\generatorname{} should generate an image that aligns with a held-out description of the expected inference.
To improve the~\generatorname{}, we feed the generated images to the VLM to verify whether they satisfy the inference, and optimize the weights to maximize this score.
For the VLM, we use Idefics2~\cite{laurencon2024what}, a late-fusion VLM that achieves superior performance on OK-VQA, a knowledge-centric question answering benchmark~\cite{okvqa}.
We automatically generate the questions used to check these inferences using GPT4o~\cite{oai2024gpt4o}.
We then compare our method against the base prompt and prompt expansion.
Since our method is orthogonal and complementary to prompting, we can apply the learned weights on top.
For prompt expansion we use GPT4o, following the same procedure as DALL-E 3~\cite{BetkerImprovingIG}. We compare performance following the same protocol as~\citet{fu2024commonsenseti}, which uses GPT4o to provide binary (yes or no) judgements of image correctness.

\noindent\textbf{Our method improves on top of prompt expansion.} We report quantitative results in~\autoref{tab:ct2i}.
Both Flux Schnell and Dev perform poorly when prompted with the base prompt, with an average accuracy of 24\% or less, demonstrating that the model inherently lacks commonsense understanding.
Our method, when applied on top, improves performance by at least 20\%.
There are also many cases where prompt expansion is insufficient; our method improves prompt expansion by at least 11\%.
In these cases, the model completely fails to visually represent the phenomenon, regardless of the amount of explanation.
This might be because the phenomenon is uncommon and underrepresented in the training data, and explicitly teaching the model with feedback is helpful.

\noindent\textbf{Commonsense needs verification over suggestion.} We show qualitative examples in~\autoref{fig:ct2i}.
The base prompt simply generates the primary object while ignoring other information that imply subtle inferences.
Sometimes the expanded prompt introduces conflicting information (top half of~\autoref{fig:ct2i}).
Although we use GPT4o for both expansion and question generation, the model can produce better inferences when instructed to check for correctness rather than simply adding more detail.
Finally, even if the expanded prompt explicitly includes the correct inference, the~\generatorname{} can still fail (bottom half of~\autoref{fig:ct2i}).
Since our method is verification centric, it is better at enforcing the inference rather than leaving it as a suggestion.

\subsection{Visual Prompting}\label{subsec:visual_prompting}
\textbf{Setup.} The default interface of~\generatorname{}s is restricted to text, due to its reliance on static text embeddings~\cite{2020t5,radford2021learning}. 
In contrast, VLMs can process arbitrary multimodal inputs, including visual prompts~\cite{cai2023vipllava} where the instruction is jointly defined in image and text.
This multimodal instruction is more conceptual, for example pointing at a spatial location and annotating what should be there.
Unlike spatial control~\cite{zhang2023adding}, where the~\generatorname{} should exactly copy the structure of the input image,
in visual prompting the model needs to reason about what the instruction is asking for.
To implement control with visual prompts, we overlay the image instruction on top of the generated image, which is jointly fed with the text instruction to the VLM.
We use Qwen2.5-VL~\cite{Qwen2.5-VL} for our VLM, as it is specifically trained on data related to spatial understanding and object grounding.
Next, we will discuss some specific use cases that can be implemented as visual prompts.

\noindent\textbf{Color Palette.} Here, our objective is to produce an image that adheres to a given color palette.
To achieve this, we overlay the palette at the bottom of the generated image and query the VLM to verify that the colors match.
We display qualitative examples in~\autoref{fig:palette}.
We control with three distinct color palettes, which we call \textit{dark academia}, \textit{pastel}, and \textit{retro}.
Evidently, our method is able to faithfully replicate the colors on naturalistic and artistic scenes with varying lighting conditions and 3D effects.

\noindent\textbf{Line Weight.} We now explore using our method to control line weight in cartoons.
We draw a red line at the bottom of the generated image and ask the VLM if the line is representative of the image's line weight.
In~\autoref{fig:line_weight} we show qualitative examples, where our method can make the line weight thinner or thicker based on the overlaid line.
Through visual prompting, we can create visual abstractions and bind different meanings based on the question.
In this case we use lines to denote the desired edge thickness, whereas in the next section we use them to denote spatial position.

\noindent\textbf{Horizon Position.} Lines can also be used to control the position of the horizon.
We overlay a red line on the generated image and ask the VLM to check if the horizon of the scene is aligned.
We show qualitative examples in~\autoref{fig:horizon}, where our method is able to raise and lower the horizon according to the image instruction.
These examples make it clear why overlaying the image instruction is helpful; one can check the spatial alignment
simply by assessing the distance between the actual horizon and the red line.

\noindent\textbf{Relative Depth.} We now show how points can be used to specify relative depth.
We place two red points, labeled as Point A and Point B, and use the VLM to verify which point is in the foreground vs. background or which point is closer vs. further from the camera.
As seen in~\autoref{fig:depth}, our method is able to move around two subjects to enforce the control.
In the first case, our method shrinks the dog to ensure that Point A lies on the background rather than the dog.
The second case imposes the reverse control, where Point A should now lie on the foreground and Point B on the background.
Our method instead enlarges the dog and creates empty space to its right to satisfy the visual prompt.
Finally, in the last case our method is able to adjust the depth ordering of the two subjects, where both animals start at roughly the same distance from the camera then the dog is moved in front of the sheep.
These examples also demonstrate how our method is able to follow truly multimodal instructions,
where the VLM needs to distinguish between the two points by associating the red labels on the image instruction
and the references to the labels in the question.

\noindent\textbf{Visual Composition.} We now show how visual prompting can be used to control composition with abstract art.
This setting is inspired by the work of Piet Mondrian, who is best known for his abstract De Stijl paintings comprised of simple primary colors and lines.
Mondrian paintings can be interpreted as highly abstracted Dutch landscapes, with elements like trees and the horizon reduced to vertical and horizontal lines~\cite{mondrianwiki}.
In~\autoref{fig:mondrian} we show how our method can be used to perform the inverse problem and produce images that match the structure of these paintings.
To make the task more clear, we annotate the paintings with additional instructions in red.
We point to specific elements and instruct that \textit{``this line represents vertical object like a tree''} or \textit{``this region is a large expanse like a field.''}
Our method is able to follow these instructions and produce images with the desired composition,
such as a wheat field in the boxed region or a lighthouse aligned with a vertical line.
Note that there exists no paired data of abstract to landscape paintings,
yet this transformation is possible when framed as a discriminative comparison task with our method.

\begin{figure}[t!]
  \centering
  \includegraphics[width=\linewidth]{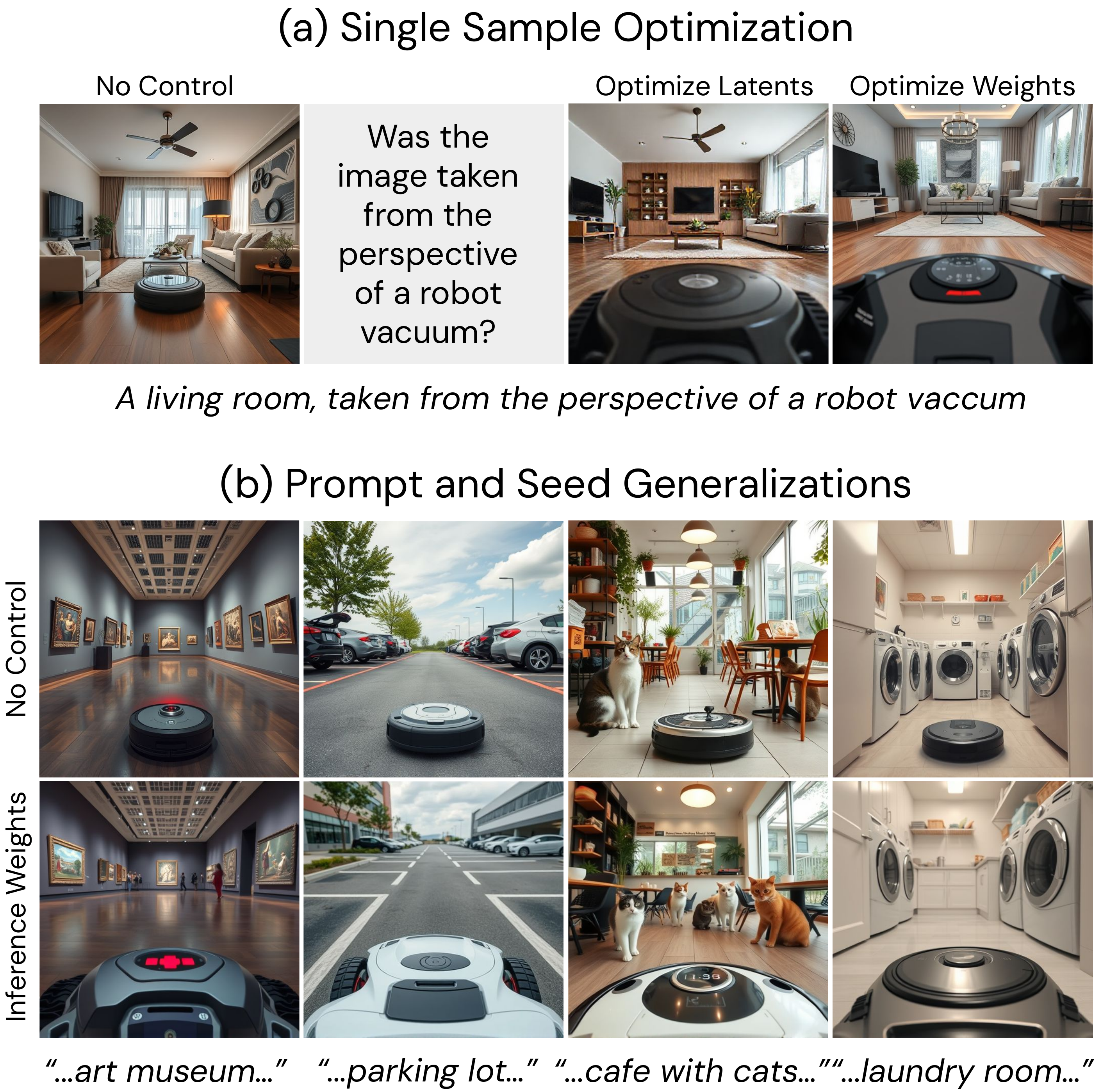}
  \caption{\textbf{Weight Generalizations.} We show that latent and weight optimization can achieve comparable results in a controlled single sample optimization setting in (a). We then show that these same weights can generalize to new seeds and prompts in a single inference pass, whereas latent optimization would need to be re-run with significant computational cost, in (b).}
  \label{fig:guidance_vs_lora}
\end{figure}

\subsection{Optimizing Latents vs. Weights}\label{subsec:latent_vs_weights}
\textbf{Setup.} Although latent optimization may seem significantly cheaper than weight optimization, we show that this is not the case. 
We compare both approaches for guiding image generation, using our VLM loss (see~\autoref{eq:vlm_loss}). 
Unlike the previous experiments, here we follow the setup of~\citet{eyring2024reno}, where the goal is to optimize a single seed such that the final image adheres to the input control.
While this setup is originally designed to optimize latents, we can use the same gradients to instead optimize model weights, with near-identical implementations.
For our VLM, we use Idefics2~\cite{laurencon2024what}.

\noindent \textbf{Computational Cost Comparison.} In~\autoref{tab:efficiency} we compare the computational cost
of latent and weight optimization. 
The gradient computation through the VLM, with respect to the clean image estimate $\hat{x}_0$, is identical for both methods. The cost slightly differs for the computation through the image generator, or the gradient of the clean image estimate $\hat{x}_0$ with respect to the input noise $z_t$.
In this case, optimizing the weights increases the VRAM usage by 5\%. 
However, weight optimization also benefits from compute amortization. 
After one minute of optimization upfront, every subsequent application of the weights is more than 200x faster.

\begin{table}[t!]
    \centering
    \begin{tabular}{l|cc}
        \toprule
        Metric & Latents & Weights \\
        \midrule
        VLM VRAM & 21GB & 21GB\\
        Image Generator VRAM & 42GB & 44GB\\
        Optimization Time & 56s & 66s\\
        Inference Time & 56s & 0.3s\\
        \bottomrule
    \end{tabular}
    \caption{\textbf{Computational Cost Comparison.} We compare the memory and runtime cost of latent and weight optimization, averaged over 100 runs.}
    \label{tab:efficiency}
\end{table}

\noindent \textbf{Weight Generalizations.} In~\autoref{fig:guidance_vs_lora}, we show an example of controlling with an unusual visual perspective.
We optimize latents and weights with the same gradients, on a single seed and single prompt, and they yield similar results (\autoref{fig:guidance_vs_lora}a).
However, these same weights can also generalize to unseen seeds and prompts, unlike latent optimization.
The optimized weights can apply the same perspective control to new scenes in a single inference pass, making the camera appear lower to the ground and raising the horizon (\autoref{fig:guidance_vs_lora}b).
Note that this single-sample setting for weight optimization is unusual.
Traditional fine-tuning methods typically optimize on multiple seeds and reference images~\cite{ruiz2023dreambooth},
whereas we show it is possible to optimize the weights in an extremely data-limited setting on a single seed without any reference images.

\section{\label{sec:discussion} Limitations}

\begin{figure}[t!]
  \centering
  \includegraphics[width=\linewidth]{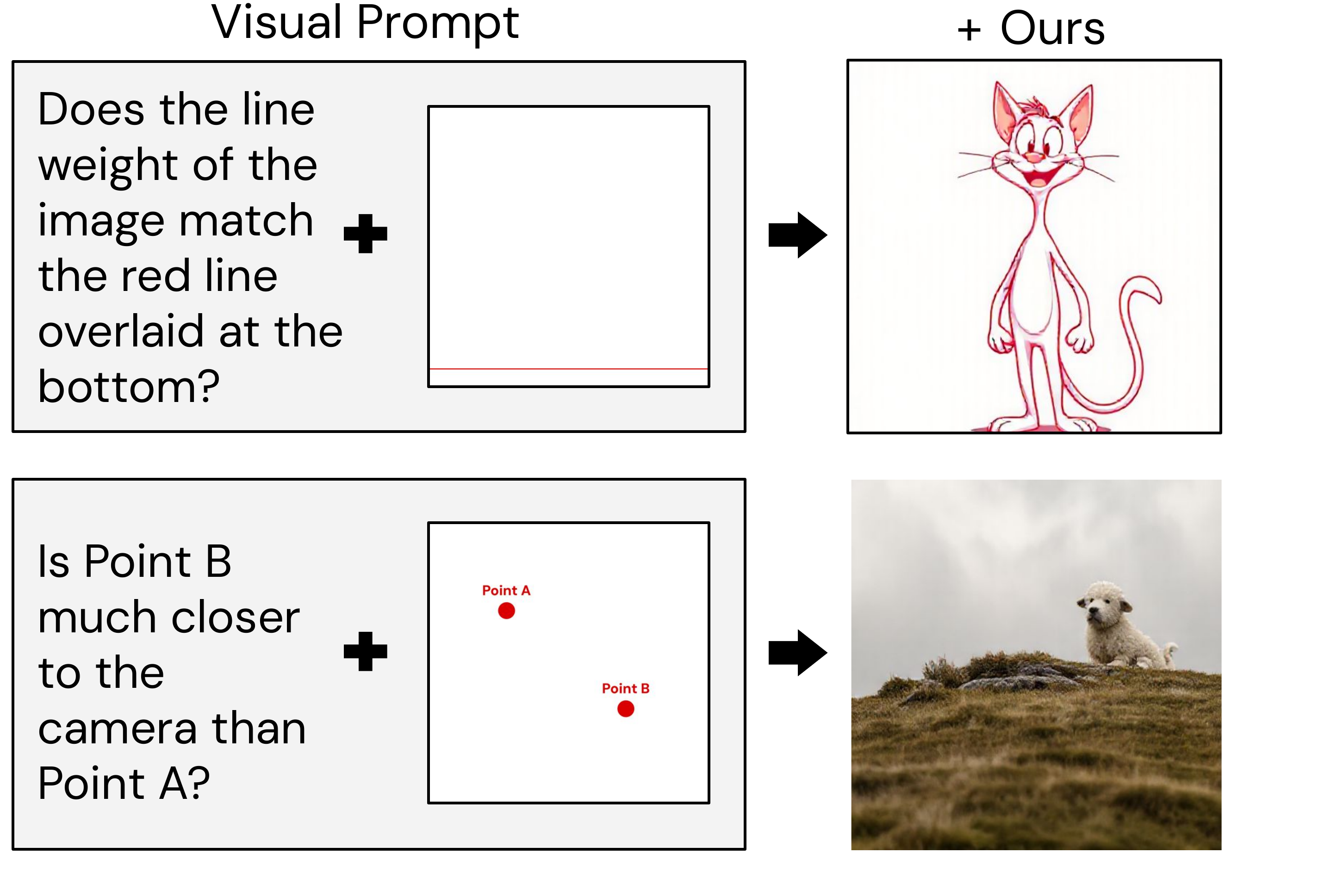}
  \caption{\textbf{Limitations.} Our method can misinterpret the visual prompt, for example changing line color when only line weight is specified or aggressively raising the ground level to satisfy a depth constraint.}
  \label{fig:limitations}
\end{figure}

Since our method relies on feedback from a model, it is susceptible to reward hacking~\cite{clark2016faulty,pan2022rewardhacking,pmlr-v202-gao23h},
where the VLM exploits ambiguities in the task specification to achieve a high reward.
In~\autoref{fig:limitations}, we show two examples of such misinterpretations.
When a red line is used to specify line width, our method may simultaneously change the line color from black to red. 
When the prompt is ambiguous, for example whether the points are bound to subjects or other elements in the scene, the model may produce an unintended but technically correct result. However, this same phenomenon could also be a useful tool for understanding how VLMs behave. One could use our method to mine adversarial images~\cite{hendrycks2021nae} where VLMs differ from human judgements; for example, cases where VLMs are blind to some visual property or are over-relying on certain visual cues. 
These automatically generated images could then be used as additional training data to improve the VLM's robustness.

\section{\label{sec:conclusion} Conclusion}
We proposed a method that distills the deliberative ability of VLMs into image generators.
Our method improves accuracy on challenging image generation tasks that require commonsense understanding, and provides new ways to control models through multimodal visual prompting.
Our method is designed to be broadly applicable, and can utilize arbitrary off-the-shelf VLMs, selected based on their strengths for a given task. As open-source VLMs scale and improve in visual understanding and instruction following, we anticipate that our method can be used as a flexible and general framework for improving image generation.

\section{Acknowledgements}
We would like to thank Anastasis Germanidis, Yining Shi, Alexander Pan, Chung Min Kim, Boyi Li, and David Chan for helpful discussions and feedback. We would also like to thank the folks at Stochastic Labs, especially Vero Bollow, Alexander Reben, and Joel Simon, for previewing early prototypes of this work.
{
    \small
    \bibliographystyle{ieeenat_fullname}
    \bibliography{main}

\begin{thebibliography}{85}
\providecommand{\natexlab}[1]{#1}
\providecommand{\url}[1]{\texttt{#1}}
\expandafter\ifx\csname urlstyle\endcsname\relax
  \providecommand{\doi}[1]{doi: #1}\else
  \providecommand{\doi}{doi: \begingroup \urlstyle{rm}\Url}\fi

\bibitem[Agrawal et~al.(2015)Agrawal, Lu, Antol, Mitchell, Zitnick, Batra, and Parikh]{agrawal2015vqa}
Aishwarya Agrawal, Jiasen Lu, Stanislaw Antol, Margaret Mitchell, C.~Lawrence Zitnick, Dhruv Batra, and Devi Parikh.
\newblock Vqa: Visual question answering.
\newblock 2015.

\bibitem[Alayrac et~al.(2022)Alayrac, Donahue, Luc, Miech, Barr, Hasson, Lenc, Mensch, Millican, Reynolds, Ring, Rutherford, Cabi, Han, Gong, Samangooei, Monteiro, Menick, Borgeaud, Brock, Nematzadeh, Sharifzadeh, Binkowski, Barreira, Vinyals, Zisserman, and Simonyan]{alayrac2022flamingo}
Jean-Baptiste Alayrac, Jeff Donahue, Pauline Luc, Antoine Miech, Iain Barr, Yana Hasson, Karel Lenc, Arthur Mensch, Katherine Millican, Malcolm Reynolds, Roman Ring, Eliza Rutherford, Serkan Cabi, Tengda Han, Zhitao Gong, Sina Samangooei, Marianne Monteiro, Jacob Menick, Sebastian Borgeaud, Andrew Brock, Aida Nematzadeh, Sahand Sharifzadeh, Mikolaj Binkowski, Ricardo Barreira, Oriol Vinyals, Andrew Zisserman, and Karen Simonyan.
\newblock Flamingo: a visual language model for few-shot learning.
\newblock In \emph{Advances in Neural Information Processing Systems}, 2022.

\bibitem[Albergo and Vanden-Eijnden(2023)]{albergo2023building}
Michael~Samuel Albergo and Eric Vanden-Eijnden.
\newblock Building normalizing flows with stochastic interpolants.
\newblock In \emph{The Eleventh International Conference on Learning Representations}, 2023.

\bibitem[Bai et~al.(2025)Bai, Chen, Liu, Wang, Ge, Song, Dang, Wang, Wang, Tang, Zhong, Zhu, Yang, Li, Wan, Wang, Ding, Fu, Xu, Ye, Zhang, Xie, Cheng, Zhang, Yang, Xu, and Lin]{Qwen2.5-VL}
Shuai Bai, Keqin Chen, Xuejing Liu, Jialin Wang, Wenbin Ge, Sibo Song, Kai Dang, Peng Wang, Shijie Wang, Jun Tang, Humen Zhong, Yuanzhi Zhu, Mingkun Yang, Zhaohai Li, Jianqiang Wan, Pengfei Wang, Wei Ding, Zheren Fu, Yiheng Xu, Jiabo Ye, Xi Zhang, Tianbao Xie, Zesen Cheng, Hang Zhang, Zhibo Yang, Haiyang Xu, and Junyang Lin.
\newblock Qwen2.5-vl technical report.
\newblock \emph{arXiv preprint arXiv:2502.13923}, 2025.

\bibitem[Bansal et~al.(2024)Bansal, Chu, Schwarzschild, Sengupta, Goldblum, Geiping, and Goldstein]{bansal2024universal}
Arpit Bansal, Hong-Min Chu, Avi Schwarzschild, Roni Sengupta, Micah Goldblum, Jonas Geiping, and Tom Goldstein.
\newblock Universal guidance for diffusion models.
\newblock In \emph{The Twelfth International Conference on Learning Representations}, 2024.

\bibitem[Betker et~al.()Betker, Goh, Jing, TimBrooks, Wang, Li, LongOuyang, JuntangZhuang, JoyceLee, YufeiGuo, WesamManassra, PrafullaDhariwal, CaseyChu, YunxinJiao, and Ramesh]{BetkerImprovingIG}
James Betker, Gabriel Goh, Li Jing, † TimBrooks, Jianfeng Wang, Linjie Li, † LongOuyang, † JuntangZhuang, † JoyceLee, † YufeiGuo, † WesamManassra, † PrafullaDhariwal, † CaseyChu, † YunxinJiao, and Aditya Ramesh.
\newblock Improving image generation with better captions.

\bibitem[Black et~al.(2024)Black, Janner, Du, Kostrikov, and Levine]{black2024training}
Kevin Black, Michael Janner, Yilun Du, Ilya Kostrikov, and Sergey Levine.
\newblock Training diffusion models with reinforcement learning.
\newblock In \emph{The Twelfth International Conference on Learning Representations}, 2024.

\bibitem[Brown et~al.(2025)Brown, Juravsky, Ehrlich, Clark, Le, Re, and Mirhoseini]{brown2025large}
Bradley Brown, Jordan Juravsky, Ryan~Saul Ehrlich, Ronald Clark, Quoc~V Le, Christopher Re, and Azalia Mirhoseini.
\newblock Large language monkeys: Scaling inference compute with repeated sampling, 2025.

\bibitem[Brown et~al.(2020)Brown, Mann, Ryder, Subbiah, Kaplan, Dhariwal, Neelakantan, Shyam, Sastry, Askell, et~al.]{brown2020language}
Tom Brown, Benjamin Mann, Nick Ryder, Melanie Subbiah, Jared~D Kaplan, Prafulla Dhariwal, Arvind Neelakantan, Pranav Shyam, Girish Sastry, Amanda Askell, et~al.
\newblock Language models are few-shot learners.
\newblock \emph{Advances in neural information processing systems}, 33:\penalty0 1877--1901, 2020.

\bibitem[Cai et~al.(2024{\natexlab{a}})Cai, Liu, Mustikovela, Meyer, Chai, Park, and Lee]{cai2023vipllava}
Mu Cai, Haotian Liu, Siva~Karthik Mustikovela, Gregory~P. Meyer, Yuning Chai, Dennis Park, and Yong~Jae Lee.
\newblock Making large multimodal models understand arbitrary visual prompts.
\newblock In \emph{IEEE Conference on Computer Vision and Pattern Recognition}, 2024{\natexlab{a}}.

\bibitem[Cai et~al.(2024{\natexlab{b}})Cai, Liu, Mustikovela, Meyer, Chai, Park, and Lee]{cai2024vip}
Mu Cai, Haotian Liu, Siva~Karthik Mustikovela, Gregory~P Meyer, Yuning Chai, Dennis Park, and Yong~Jae Lee.
\newblock Vip-llava: Making large multimodal models understand arbitrary visual prompts.
\newblock In \emph{Proceedings of the IEEE/CVF Conference on Computer Vision and Pattern Recognition}, pages 12914--12923, 2024{\natexlab{b}}.

\bibitem[Chen et~al.(2025)Chen, Wu, Liu, Pan, Liu, Xie, Yu, and Ruan]{chen2025janus}
Xiaokang Chen, Zhiyu Wu, Xingchao Liu, Zizheng Pan, Wen Liu, Zhenda Xie, Xingkai Yu, and Chong Ruan.
\newblock Janus-pro: Unified multimodal understanding and generation with data and model scaling.
\newblock \emph{arXiv preprint arXiv:2501.17811}, 2025.

\bibitem[Cho et~al.(2023)Cho, Zala, and Bansal]{Cho2023DallEval}
Jaemin Cho, Abhay Zala, and Mohit Bansal.
\newblock Dall-eval: Probing the reasoning skills and social biases of text-to-image generation models.
\newblock In \emph{ICCV}, 2023.

\bibitem[Cho et~al.(2024)Cho, Hu, Garg, Anderson, Krishna, Baldridge, Bansal, Pont-Tuset, and Wang]{JaeminCho2024}
Jaemin Cho, Yushi Hu, Roopal Garg, Peter Anderson, Ranjay Krishna, Jason Baldridge, Mohit Bansal, Jordi Pont-Tuset, and Su Wang.
\newblock {Davidsonian Scene Graph: Improving Reliability in Fine-Grained Evaluation for Text-to-Image Generation}.
\newblock In \emph{ICLR}, 2024.

\bibitem[Deng et~al.(2009)Deng, Dong, Socher, Li, Li, and Fei-Fei]{imagenet}
Jia Deng, Wei Dong, Richard Socher, Li-Jia Li, Kai Li, and Li Fei-Fei.
\newblock Imagenet: A large-scale hierarchical image database.
\newblock In \emph{2009 IEEE Conference on Computer Vision and Pattern Recognition}, 2009.

\bibitem[Dhariwal and Nichol(2021)]{dhariwal2021}
Prafulla Dhariwal and Alex Nichol.
\newblock Diffusion models beat gans on image synthesis.
\newblock In \emph{Proceedings of the 35th International Conference on Neural Information Processing Systems}, Red Hook, NY, USA, 2021. Curran Associates Inc.

\bibitem[Donahue et~al.(2017)Donahue, Hendricks, Rohrbach, Venugopalan, Guadarrama, Saenko, and Darrell]{donahue2017lrcn}
Jeff Donahue, Lisa~Anne Hendricks, Marcus Rohrbach, Subhashini Venugopalan, Sergio Guadarrama, Kate Saenko, and Trevor Darrell.
\newblock Long-term recurrent convolutional networks for visual recognition and description.
\newblock \emph{IEEE Transactions on Pattern Analysis and Machine Intelligence}, 39\penalty0 (4):\penalty0 677--691, 2017.

\bibitem[Epstein et~al.(2023)Epstein, Jabri, Poole, Efros, and Holynski]{epstein2023selfguidance}
Dave Epstein, Allan Jabri, Ben Poole, Alexei~A. Efros, and Aleksander Holynski.
\newblock Diffusion self-guidance for controllable image generation.
\newblock 2023.

\bibitem[Esser et~al.(2024)Esser, Kulal, Blattmann, Entezari, M{\"u}ller, Saini, Levi, Lorenz, Sauer, Boesel, Podell, Dockhorn, English, and Rombach]{esser2024scaling}
Patrick Esser, Sumith Kulal, Andreas Blattmann, Rahim Entezari, Jonas M{\"u}ller, Harry Saini, Yam Levi, Dominik Lorenz, Axel Sauer, Frederic Boesel, Dustin Podell, Tim Dockhorn, Zion English, and Robin Rombach.
\newblock Scaling rectified flow transformers for high-resolution image synthesis.
\newblock In \emph{Forty-first International Conference on Machine Learning}, 2024.

\bibitem[Eyring et~al.(2024)Eyring, Karthik, Roth, Dosovitskiy, and Akata]{eyring2024reno}
Luca Eyring, Shyamgopal Karthik, Karsten Roth, Alexey Dosovitskiy, and Zeynep Akata.
\newblock Reno: Enhancing one-step text-to-image models through reward-based noise optimization.
\newblock \emph{Neural Information Processing Systems (NeurIPS)}, 2024.

\bibitem[Fan et~al.(2023)Fan, Watkins, Du, Liu, Ryu, Boutilier, Abbeel, Ghavamzadeh, Lee, and Lee]{fan2023dpok}
Ying Fan, Olivia Watkins, Yuqing Du, Hao Liu, Moonkyung Ryu, Craig Boutilier, Pieter Abbeel, Mohammad Ghavamzadeh, Kangwook Lee, and Kimin Lee.
\newblock Dpok: reinforcement learning for fine-tuning text-to-image diffusion models.
\newblock In \emph{Proceedings of the 37th International Conference on Neural Information Processing Systems}, Red Hook, NY, USA, 2023. Curran Associates Inc.

\bibitem[Fu et~al.(2024)Fu, He, Lu, Wang, and Roth]{fu2024commonsenseti}
Xingyu Fu, Muyu He, Yujie Lu, William~Yang Wang, and Dan Roth.
\newblock Commonsense-t2i challenge: Can text-to-image generation models understand commonsense?
\newblock In \emph{First Conference on Language Modeling}, 2024.

\bibitem[Gandikota et~al.(2023)Gandikota, Materzy\'nska, Zhou, Torralba, and Bau]{gandikota2023sliders}
Rohit Gandikota, Joanna Materzy\'nska, Tingrui Zhou, Antonio Torralba, and David Bau.
\newblock Concept sliders: Lora adaptors for precise control in diffusion models.
\newblock \emph{arXiv preprint arXiv:2311.12092}, 2023.

\bibitem[Gao et~al.(2023)Gao, Schulman, and Hilton]{pmlr-v202-gao23h}
Leo Gao, John Schulman, and Jacob Hilton.
\newblock Scaling laws for reward model overoptimization.
\newblock In \emph{Proceedings of the 40th International Conference on Machine Learning}, pages 10835--10866. PMLR, 2023.

\bibitem[Gao et~al.(2024)Gao, Hoogeboom, Heek, Bortoli, Murphy, and Salimans]{gao2025diffusionmeetsflow}
Ruiqi Gao, Emiel Hoogeboom, Jonathan Heek, Valentin~De Bortoli, Kevin~P. Murphy, and Tim Salimans.
\newblock Diffusion meets flow matching: Two sides of the same coin.
\newblock 2024.

\bibitem[Hendrycks et~al.(2021)Hendrycks, Zhao, Basart, Steinhardt, and Song]{hendrycks2021nae}
Dan Hendrycks, Kevin Zhao, Steven Basart, Jacob Steinhardt, and Dawn Song.
\newblock Natural adversarial examples.
\newblock \emph{CVPR}, 2021.

\bibitem[Heusel et~al.(2017)Heusel, Ramsauer, Unterthiner, Nessler, and Hochreiter]{heusel2017gans}
Martin Heusel, Hubert Ramsauer, Thomas Unterthiner, Bernhard Nessler, and Sepp Hochreiter.
\newblock Gans trained by a two time-scale update rule converge to a local nash equilibrium.
\newblock \emph{Advances in neural information processing systems}, 30, 2017.

\bibitem[Ho et~al.(2020)Ho, Jain, and Abbeel]{ho2020denoising}
Jonathan Ho, Ajay Jain, and Pieter Abbeel.
\newblock Denoising diffusion probabilistic models.
\newblock In \emph{Advances in Neural Information Processing Systems}, pages 6840--6851, 2020.

\bibitem[Hu et~al.(2022)Hu, Shen, Wallis, Allen-Zhu, Li, Wang, Wang, and Chen]{hu2022lora}
Edward~J Hu, Yelong Shen, Phillip Wallis, Zeyuan Allen-Zhu, Yuanzhi Li, Shean Wang, Lu Wang, and Weizhu Chen.
\newblock Lo{RA}: Low-rank adaptation of large language models.
\newblock In \emph{International Conference on Learning Representations}, 2022.

\bibitem[Jack~Clark(2016)]{clark2016faulty}
Dario~Amodei Jack~Clark.
\newblock Faulty reward functions in the wild.
\newblock \url{https://openai.com/index/faulty-reward-functions}, 2016.

\bibitem[Jayasumana et~al.(2024)Jayasumana, Ramalingam, Veit, Glasner, Chakrabarti, and Kumar]{jayasumana2024rethinking}
Sadeep Jayasumana, Srikumar Ramalingam, Andreas Veit, Daniel Glasner, Ayan Chakrabarti, and Sanjiv Kumar.
\newblock Rethinking fid: Towards a better evaluation metric for image generation.
\newblock In \emph{Proceedings of the IEEE/CVF Conference on Computer Vision and Pattern Recognition}, pages 9307--9315, 2024.

\bibitem[Jin et~al.(2023)Jin, Zhang, Hold-Geoffroy, Wang, Matzen, Sticha, and Fouhey]{jin2022PerspectiveFields}
Linyi Jin, Jianming Zhang, Yannick Hold-Geoffroy, Oliver Wang, Kevin Matzen, Matthew Sticha, and David~F. Fouhey.
\newblock Perspective fields for single image camera calibration.
\newblock \emph{CVPR}, 2023.

\bibitem[Kahneman(2011)]{kahneman2011thinking}
Daniel Kahneman.
\newblock \emph{Thinking, Fast and Slow}.
\newblock Farrar, Straus and Giroux, 2011.

\bibitem[Kim et~al.(2022)Kim, Jang, Lee, Hong, Seo, and Kim]{kim2022dag}
Gyeongnyeon Kim, Wooseok Jang, Gyuseong Lee, Susung Hong, Junyoung Seo, and Seungryong Kim.
\newblock Dag: Depth-aware guidance with denoising diffusion probabilistic models.
\newblock \emph{arXiv preprint arXiv: Arxiv-2212.08861}, 2022.

\bibitem[Kingma and Ba(2015)]{DBLP:journals/corr/KingmaB14}
Diederik~P. Kingma and Jimmy Ba.
\newblock Adam: {A} method for stochastic optimization.
\newblock In \emph{3rd International Conference on Learning Representations, {ICLR} 2015, San Diego, CA, USA, May 7-9, 2015, Conference Track Proceedings}, 2015.

\bibitem[Ku et~al.(2024)Ku, Jiang, Wei, Yue, and Chen]{ku2024viescoreexplainablemetricsconditional}
Max Ku, Dongfu Jiang, Cong Wei, Xiang Yue, and Wenhu Chen.
\newblock Viescore: Towards explainable metrics for conditional image synthesis evaluation, 2024.

\bibitem[Labs(2024)]{flux2024}
Black~Forest Labs.
\newblock Flux.
\newblock \url{https://github.com/black-forest-labs/flux}, 2024.

\bibitem[Laurencon et~al.(2024)Laurencon, Tronchon, Cord, and Sanh]{laurencon2024what}
Hugo Laurencon, Leo Tronchon, Matthieu Cord, and Victor Sanh.
\newblock What matters when building vision-language models?
\newblock In \emph{The Thirty-eighth Annual Conference on Neural Information Processing Systems}, 2024.

\bibitem[Lian et~al.(2024)Lian, Li, Yala, and Darrell]{lian2024llmgrounded}
Long Lian, Boyi Li, Adam Yala, and Trevor Darrell.
\newblock {LLM}-grounded diffusion: Enhancing prompt understanding of text-to-image diffusion models with large language models.
\newblock \emph{Transactions on Machine Learning Research}, 2024.
\newblock Featured Certification.

\bibitem[Lin et~al.(2014)Lin, Maire, Belongie, Hays, Perona, Ramanan, Doll{\'a}r, and Zitnick]{lin2014microsoft}
Tsung-Yi Lin, Michael Maire, Serge Belongie, James Hays, Pietro Perona, Deva Ramanan, Piotr Doll{\'a}r, and C~Lawrence Zitnick.
\newblock Microsoft coco: Common objects in context.
\newblock In \emph{Computer vision--ECCV 2014: 13th European conference, zurich, Switzerland, September 6-12, 2014, proceedings, part v 13}, pages 740--755. Springer, 2014.

\bibitem[Lin et~al.(2024)Lin, Pathak, Li, Li, Xia, Neubig, Zhang, and Ramanan]{lin2024vqa}
Zhiqiu Lin, Deepak Pathak, Baiqi Li, Jiayao Li, Xide Xia, Graham Neubig, Pengchuan Zhang, and Deva Ramanan.
\newblock Evaluating text-to-visual generation with image-to-text generation.
\newblock In \emph{Computer Vision – ECCV 2024: 18th European Conference, Milan, Italy, September 29–October 4, 2024, Proceedings, Part IX}, page 366–384, Berlin, Heidelberg, 2024. Springer-Verlag.

\bibitem[Lipman et~al.(2023)Lipman, Chen, Ben-Hamu, Nickel, and Le]{lipman2023flow}
Yaron Lipman, Ricky T.~Q. Chen, Heli Ben-Hamu, Maximilian Nickel, and Matthew Le.
\newblock Flow matching for generative modeling.
\newblock In \emph{The Eleventh International Conference on Learning Representations}, 2023.

\bibitem[Liu et~al.(2023{\natexlab{a}})Liu, Li, Wu, and Lee]{liu2023visual}
Haotian Liu, Chunyuan Li, Qingyang Wu, and Yong~Jae Lee.
\newblock Visual instruction tuning.
\newblock In \emph{Thirty-seventh Conference on Neural Information Processing Systems}, 2023{\natexlab{a}}.

\bibitem[Liu(2024)]{liu2024dalleprompt}
Tianyang Liu.
\newblock System prompt of chatgpt.
\newblock \url{https://leoii22.com/blog/2024/system-prompt-of-gpt4/}, 2024.

\bibitem[Liu et~al.(2023{\natexlab{b}})Liu, Gong, and qiang liu]{liu2023flow}
Xingchao Liu, Chengyue Gong, and qiang liu.
\newblock Flow straight and fast: Learning to generate and transfer data with rectified flow.
\newblock In \emph{The Eleventh International Conference on Learning Representations}, 2023{\natexlab{b}}.

\bibitem[Luo et~al.(2024)Luo, Darrell, Wang, Goldman, and Holynski]{luo2024readoutguidance}
Grace Luo, Trevor Darrell, Oliver Wang, Dan~B Goldman, and Aleksander Holynski.
\newblock Readout guidance: Learning control from diffusion features.
\newblock 2024.

\bibitem[Luo et~al.(2023)Luo, Tan, Huang, Li, and Zhao]{luo2023latent}
Simian Luo, Yiqin Tan, Longbo Huang, Jian Li, and Hang Zhao.
\newblock Latent consistency models: Synthesizing high-resolution images with few-step inference, 2023.

\bibitem[Ma et~al.(2025)Ma, Tong, Jia, Hu, Su, Zhang, Yang, Li, Jaakkola, Jia, and Xie]{ma2025inferencetimescaling}
Nanye Ma, Shangyuan Tong, Haolin Jia, Hexiang Hu, Yu-Chuan Su, Mingda Zhang, Xuan Yang, Yandong Li, Tommi Jaakkola, Xuhui Jia, and Saining Xie.
\newblock Inference-time scaling for diffusion models beyond scaling denoising steps, 2025.

\bibitem[Marino et~al.(2019)Marino, Rastegari, Farhadi, and Mottaghi]{okvqa}
Kenneth Marino, Mohammad Rastegari, Ali Farhadi, and Roozbeh Mottaghi.
\newblock Ok-vqa: A visual question answering benchmark requiring external knowledge.
\newblock In \emph{Conference on Computer Vision and Pattern Recognition (CVPR)}, 2019.

\bibitem[Mondrian(1929)]{Mondrian1929}
Piet Mondrian.
\newblock Composition, 1929.

\bibitem[Mondrian(1930)]{Mondrian1930}
Piet Mondrian.
\newblock Composition with red, blue, and yellow, 1930.

\bibitem[Nguyen et~al.(2017)Nguyen, Clune, Bengio, Dosovitskiy, and Yosinski]{nguyen2017plug}
Anh Nguyen, Jeff Clune, Yoshua Bengio, Alexey Dosovitskiy, and Jason Yosinski.
\newblock Plug and play generative networks: Conditional iterative generation of images in latent space.
\newblock In \emph{CVPR}, 2017.

\bibitem[OpenAI(2024)]{oai2024gpt4o}
OpenAI.
\newblock Gpt-4o system card, 2024.

\bibitem[OpenAI(2025)]{oai2025gpt4oimagegen}
OpenAI.
\newblock Introducing 4o image generation, 2025.

\bibitem[Pan et~al.(2022)Pan, Bhatia, and Steinhardt]{pan2022rewardhacking}
Alexander Pan, Kush Bhatia, and Jacob Steinhardt.
\newblock The effects of reward misspecification: Mapping and mitigating misaligned models.
\newblock In \emph{International Conference on Learning Representations}, 2022.

\bibitem[Parmar et~al.(2022)Parmar, Zhang, and Zhu]{parmar2021cleanfid}
Gaurav Parmar, Richard Zhang, and Jun-Yan Zhu.
\newblock On aliased resizing and surprising subtleties in gan evaluation.
\newblock In \emph{CVPR}, 2022.

\bibitem[Poole et~al.(2023)Poole, Jain, Barron, and Mildenhall]{poole2023dreamfusion}
Ben Poole, Ajay Jain, Jonathan~T. Barron, and Ben Mildenhall.
\newblock Dreamfusion: Text-to-3d using 2d diffusion.
\newblock In \emph{The Eleventh International Conference on Learning Representations}, 2023.

\bibitem[Radford et~al.(2021{\natexlab{a}})Radford, Kim, Hallacy, Ramesh, Goh, Agarwal, Sastry, Askell, Mishkin, Clark, Krueger, and Sutskever]{radford21learning}
Alec Radford, Jong~Wook Kim, Chris Hallacy, Aditya Ramesh, Gabriel Goh, Sandhini Agarwal, Girish Sastry, Amanda Askell, Pamela Mishkin, Jack Clark, Gretchen Krueger, and Ilya Sutskever.
\newblock Learning transferable visual models from natural language supervision.
\newblock In \emph{Proceedings of the 38th International Conference on Machine Learning}, pages 8748--8763. PMLR, 2021{\natexlab{a}}.

\bibitem[Radford et~al.(2021{\natexlab{b}})Radford, Kim, Hallacy, Ramesh, Goh, Agarwal, Sastry, Askell, Mishkin, Clark, et~al.]{radford2021learning}
Alec Radford, Jong~Wook Kim, Chris Hallacy, Aditya Ramesh, Gabriel Goh, Sandhini Agarwal, Girish Sastry, Amanda Askell, Pamela Mishkin, Jack Clark, et~al.
\newblock Learning transferable visual models from natural language supervision.
\newblock In \emph{International conference on machine learning}, pages 8748--8763. PmLR, 2021{\natexlab{b}}.

\bibitem[Raffel et~al.(2020)Raffel, Shazeer, Roberts, Lee, Narang, Matena, Zhou, Li, and Liu]{2020t5}
Colin Raffel, Noam Shazeer, Adam Roberts, Katherine Lee, Sharan Narang, Michael Matena, Yanqi Zhou, Wei Li, and Peter~J. Liu.
\newblock Exploring the limits of transfer learning with a unified text-to-text transformer.
\newblock \emph{Journal of Machine Learning Research}, 21\penalty0 (140):\penalty0 1--67, 2020.

\bibitem[Ruiz et~al.(2023)Ruiz, Li, Jampani, Pritch, Rubinstein, and Aberman]{ruiz2023dreambooth}
Nataniel Ruiz, Yuanzhen Li, Varun Jampani, Yael Pritch, Michael Rubinstein, and Kfir Aberman.
\newblock Dreambooth: Fine tuning text-to-image diffusion models for subject-driven generation.
\newblock In \emph{Proceedings of the IEEE/CVF conference on computer vision and pattern recognition}, pages 22500--22510, 2023.

\bibitem[Sauer et~al.(2024)Sauer, Lorenz, Blattmann, and Rombach]{sauer2024add}
Axel Sauer, Dominik Lorenz, Andreas Blattmann, and Robin Rombach.
\newblock Adversarial diffusion distillation.
\newblock 2024.

\bibitem[Schramowski et~al.(2023)Schramowski, Brack, Deiseroth, and Kersting]{schramowski2022safe}
Patrick Schramowski, Manuel Brack, Björn Deiseroth, and Kristian Kersting.
\newblock Safe latent diffusion: Mitigating inappropriate degeneration in diffusion models.
\newblock In \emph{Proceedings of the {IEEE} Conference on Computer Vision and Pattern Recognition ({CVPR})}, 2023.

\bibitem[Sclar et~al.(2024)Sclar, Choi, Tsvetkov, and Suhr]{sclar2024quantifying}
Melanie Sclar, Yejin Choi, Yulia Tsvetkov, and Alane Suhr.
\newblock Quantifying language models' sensitivity to spurious features in prompt design or: How i learned to start worrying about prompt formatting.
\newblock In \emph{ICLR}, 2024.

\bibitem[Singhal et~al.(2025)Singhal, Horvitz, Teehan, Ren, Yu, McKeown, and Ranganath]{singhal2025fksteering}
Raghav Singhal, Zachary Horvitz, Ryan Teehan, Mengye Ren, Zhou Yu, Kathleen McKeown, and Rajesh Ranganath.
\newblock A general framework for inference-time scaling and steering of diffusion models, 2025.

\bibitem[Sloman(1996)]{Sloman1996TheEC}
Steven~A. Sloman.
\newblock The empirical case for two systems of reasoning.
\newblock \emph{Psychological Bulletin}, 119:\penalty0 3--22, 1996.

\bibitem[Snell et~al.(2025)Snell, Lee, Xu, and Kumar]{snell2025scaling}
Charlie~Victor Snell, Jaehoon Lee, Kelvin Xu, and Aviral Kumar.
\newblock Scaling {LLM} test-time compute optimally can be more effective than scaling parameters for reasoning.
\newblock In \emph{The Thirteenth International Conference on Learning Representations}, 2025.

\bibitem[Sohl-Dickstein et~al.(2015)Sohl-Dickstein, Weiss, Maheswaranathan, and Ganguli]{sohl2015deep}
Jascha Sohl-Dickstein, Eric Weiss, Niru Maheswaranathan, and Surya Ganguli.
\newblock Deep unsupervised learning using nonequilibrium thermodynamics.
\newblock In \emph{Proceedings of the 32nd International Conference on Machine Learning}, pages 2256--2265, Lille, France, 2015. PMLR.

\bibitem[Song et~al.(2021)Song, Meng, and Ermon]{song2021denoising}
Jiaming Song, Chenlin Meng, and Stefano Ermon.
\newblock Denoising diffusion implicit models.
\newblock In \emph{International Conference on Learning Representations}, 2021.

\bibitem[Team(2024)]{chameleonteam2024chameleonmixedmodalearlyfusionfoundation}
Chameleon Team.
\newblock Chameleon: Mixed-modal early-fusion foundation models, 2024.

\bibitem[Team(2025)]{gemma_2025}
Gemma Team.
\newblock Gemma 3.
\newblock 2025.

\bibitem[Tong et~al.(2024)Tong, Fan, Zhu, Xiong, Chen, Sinha, Rabbat, LeCun, Xie, and Liu]{tong2024metamorphmultimodalunderstandinggeneration}
Shengbang Tong, David Fan, Jiachen Zhu, Yunyang Xiong, Xinlei Chen, Koustuv Sinha, Michael Rabbat, Yann LeCun, Saining Xie, and Zhuang Liu.
\newblock Metamorph: Multimodal understanding and generation via instruction tuning, 2024.

\bibitem[Voynov et~al.(2023)Voynov, Aberman, and Cohen-Or]{voynov2023}
Andrey Voynov, Kfir Aberman, and Daniel Cohen-Or.
\newblock Sketch-guided text-to-image diffusion models.
\newblock New York, NY, USA, 2023. Association for Computing Machinery.

\bibitem[Wallace et~al.(2024)Wallace, Dang, Rafailov, Zhou, Lou, Purushwalkam, Ermon, Xiong, Joty, and Naik]{wallace2024dpo}
Bram Wallace, Meihua Dang, Rafael Rafailov, Linqi Zhou, Aaron Lou, Senthil Purushwalkam, Stefano Ermon, Caiming Xiong, Shafiq Joty, and Nikhil Naik.
\newblock Diffusion model alignment using direct preference optimization.
\newblock In \emph{2024 IEEE/CVF Conference on Computer Vision and Pattern Recognition (CVPR)}, pages 8228--8238, 2024.

\bibitem[Wang et~al.(2023)Wang, Du, Li, Yeh, and Shakhnarovich]{wang2022sjc}
Haochen Wang, Xiaodan Du, Jiahao Li, Raymond~A. Yeh, and Greg Shakhnarovich.
\newblock Score jacobian chaining: Lifting pretrained 2d diffusion models for 3d generation.
\newblock In \emph{CVPR}, 2023.

\bibitem[Whitaker(2023)]{miduguidance}
Jonathan Whitaker, 2023.

\bibitem[{Wikipedia contributors}(2025)]{mondrianwiki}
{Wikipedia contributors}.
\newblock Piet mondrian --- {Wikipedia}{,} the free encyclopedia, 2025.
\newblock [Online; accessed 7-March-2025].

\bibitem[Wu et~al.(2024)Wu, Lian, Gonzalez, Li, and Darrell]{wu2024sld}
Tsung-Han Wu, Long Lian, Joseph~E Gonzalez, Boyi Li, and Trevor Darrell.
\newblock Self-correcting llm-controlled diffusion models.
\newblock 2024.

\bibitem[Wu et~al.(2025)Wu, Sun, Li, Welleck, and Yang]{wu2025inference}
Yangzhen Wu, Zhiqing Sun, Shanda Li, Sean Welleck, and Yiming Yang.
\newblock Inference scaling laws: An empirical analysis of compute-optimal inference for {LLM} problem-solving.
\newblock In \emph{The Thirteenth International Conference on Learning Representations}, 2025.

\bibitem[Xie et~al.(2024)Xie, Chen, Chen, Cai, Tang, Lin, Zhang, Li, Zhu, Lu, and Han]{xie2024sana}
Enze Xie, Junsong Chen, Junyu Chen, Han Cai, Haotian Tang, Yujun Lin, Zhekai Zhang, Muyang Li, Ligeng Zhu, Yao Lu, and Song Han.
\newblock Sana: Efficient high-resolution image synthesis with linear diffusion transformers, 2024.

\bibitem[Xu et~al.(2023)Xu, Liu, Wu, Tong, Li, Ding, Tang, and Dong]{xu2023imagereward}
Jiazheng Xu, Xiao Liu, Yuchen Wu, Yuxuan Tong, Qinkai Li, Ming Ding, Jie Tang, and Yuxiao Dong.
\newblock Imagereward: learning and evaluating human preferences for text-to-image generation.
\newblock In \emph{Proceedings of the 37th International Conference on Neural Information Processing Systems}, pages 15903--15935, 2023.

\bibitem[Xu et~al.(2024)Xu, Huang, Cheng, Yang, Xu, Wang, Duan, Yang, Jin, Li, Teng, Yang, Zheng, Liu, Ding, Zhang, Gu, Huang, Huang, Tang, and Dong]{xu2024visionrewardfinegrainedmultidimensionalhuman}
Jiazheng Xu, Yu Huang, Jiale Cheng, Yuanming Yang, Jiajun Xu, Yuan Wang, Wenbo Duan, Shen Yang, Qunlin Jin, Shurun Li, Jiayan Teng, Zhuoyi Yang, Wendi Zheng, Xiao Liu, Ming Ding, Xiaohan Zhang, Xiaotao Gu, Shiyu Huang, Minlie Huang, Jie Tang, and Yuxiao Dong.
\newblock Visionreward: Fine-grained multi-dimensional human preference learning for image and video generation, 2024.

\bibitem[Yin et~al.(2024)Yin, Gharbi, Zhang, Shechtman, Durand, Freeman, and Park]{yin2024onestep}
Tianwei Yin, Micha{\"e}l Gharbi, Richard Zhang, Eli Shechtman, Fr{\'e}do Durand, William~T Freeman, and Taesung Park.
\newblock One-step diffusion with distribution matching distillation.
\newblock In \emph{CVPR}, 2024.

\bibitem[Zhang et~al.(2023{\natexlab{a}})Zhang, Rao, and Agrawala]{zhang2023adding}
Lvmin Zhang, Anyi Rao, and Maneesh Agrawala.
\newblock Adding conditional control to text-to-image diffusion models, 2023{\natexlab{a}}.

\bibitem[Zhang et~al.(2023{\natexlab{b}})Zhang, Lu, Wang, Yan, Yan, Qin, Wang, Yan, Wang, and Petzold]{zhang2023gpt4visiongeneralistevaluatorvisionlanguage}
Xinlu Zhang, Yujie Lu, Weizhi Wang, An Yan, Jun Yan, Lianke Qin, Heng Wang, Xifeng Yan, William~Yang Wang, and Linda~Ruth Petzold.
\newblock Gpt-4v(ision) as a generalist evaluator for vision-language tasks, 2023{\natexlab{b}}.

\end{thebibliography}
}

\clearpage
\setcounter{page}{1}

\onecolumn
\begingroup
    \centering
    \Large
    \textbf{\thetitle}\\
    \vspace{0.5em}Supplementary Material \\
    \vspace{1.0em}
\endgroup

\section{Ablations for Commonsense Inferences}
In this section, we characterize the discriminative ability of different VLMs, and its effect on our method's generation quality.\\

\noindent \textbf{VLM Scoring.} First, we compare the agreement between VLM scores and human annotations. We take 1200 images generated by Flux Schnell, which were released by CommonsenseT2I\footnote{The images can be found at \url{https://zeyofu.github.io/CommonsenseT2I/visualization\_flux\_schenel.html}.}. We have one human annotator then rate each image as correct/incorrect. We also feed these same images to different open-source VLMs, prompted with: \textit{Does the image generally fit the description ``{\small\texttt{<description>}}''? Answer with Yes or No.} We then compare the alignment of the human and VLM ratings, using the metrics of accuracy and sensitivity. To compute the classification accuracy, we compare the most likely next token with the human annotated ground-truth. For sensitivity, we extract the probabilities of the ``Yes'' and ``No'' tokens from the vocabulary distribution, and compute the difference between the correct vs. incorrect class (as labeled by the human annotations). We report the results in~\autoref{tab:vlm_scoring}, for both newer and bigger open-source VLMs, as well as the closed-source GPT4o upper bound. Evidently, newer VLMs like Idefics2 and bigger VLMs like Gemma3 27B strongly agree with human answers, with less than a 1\% difference from GPT4o. We also see that our method performs better with stronger VLMs; supervising with Idefics2 instead of LLaVA leads to a 14.5\% boost on CommonsenseT2I generations.

\begin{table}[h!]
\centering
\scriptsize
\begin{tabular}{l|ccc|ccc|c}
\toprule
& \multicolumn{3}{c|}{newer models\;\(\rightarrow\)} & \multicolumn{3}{c|}{bigger models\;\(\rightarrow\)}\\[2pt]
& LLaVA~\cite{liu2023visual} & Qwen2.5VL~\cite{Qwen2.5-VL} & Idefics2~\cite{laurencon2024what} & Gemma3 4B~\cite{gemma_2025} & Gemma3 12B & Gemma3 27B & GPT4o~\cite{oai2024gpt4o}\\
\midrule
Accuracy ($\uparrow$) & 0.7250 & 0.7892 & \textbf{0.8092} & 0.7758 & 0.7858 & \textbf{0.8108} & 0.8183\\
Sensitivity ($\uparrow$) & 0.2875 & 0.5504 & \textbf{0.5998} & 0.5536 & 0.5686 & \textbf{0.6219} & N/A\\
\bottomrule
\end{tabular}
\caption{\textbf{Newer and bigger VLMs agree more with humans on image correctness.} We report both the accuracy and sensitivity of different VLM scores compared with human annotations on CommonsenseT2I.
}
\label{tab:vlm_scoring}
\end{table}

\noindent \textbf{Automatic Question Reliability.} Next, we ablate the influence of the VLM question on the generator output. In our normal automatic question generation pipeline, we feed only the prompt to GPT4o, which correctly covers the desired commonsense inference 82\% of the time. Here, we also experiment with more ``reliable'' questions derived from the ground-truth inferences provided by CommonsenseT2I, by converting these inferences into questions with GPT4o. In~\autoref{tab:reliability} we observe that simply improving the supervision signal with more reliable questions, without modifying the fine-tuning procedure, can result in progressively larger gains in generation quality.

\begin{table}[h!]
\centering
\scriptsize
\begin{tabular}{l|ccccc}
\toprule
& \multicolumn{5}{c}{percentage of reliable questions\;\(\rightarrow\)} \\[2pt]
 & 0\% & 25\% & 50\% &  75\% & 100\% \\
\midrule
Accuracy $\% \Delta$ from No Control ($\uparrow$) & +20.1 & +23.9 & +24.8 & +27.1 & \textbf{+28.2}   \\
\bottomrule
\end{tabular}
\caption{\textbf{More reliable VLM questions lead to better generator outputs.} We ablate guiding varying proportions of samples with more reliable questions derived from the ground-truth labels from CommonsenseT2I, mixed with automatically generated questions. We report results in terms of the performance delta on top of the base prompt, or ``No Control.''
}
\label{tab:reliability}
\end{table}

\noindent \textbf{Image Quality Comparison.} Finally, we validate that our method's improvements in logical accuracy do not compromise image quality in~\autoref{tab:fid}. Following the setup from~\citet{jayasumana2024rethinking}, we use MS-COCO 30K~\cite{lin2014microsoft} as the real image set and outputs from CommonsenseT2I~\cite{fu2024commonsenseti} as the generated image set. Evidently, our method matches -- and in fact exceeds -- the image quality of the prompting baselines across the board. This improvement could be because low perceptual quality also hurts logical coherence; for example, if a water glass is malformed then it is also unlikely to be recognized as correctly ``spilling [...] onto the table.''

\begin{table*}[h!]
    \centering
    \scriptsize
    \begin{tabular}{ll|cc}
        \toprule
        Generator & Method & CMMD ($\downarrow$) & FID ($\downarrow$)\\
        \toprule
        \multirow{4}{*}{\shortstack{\textbf{Flux Schnell} \\ (single-step)}} & Base Prompt & 1.17&102.8\\
        & Base Prompt + Ours & \textbf{1.10}& \textbf{98.7}\\
        \cmidrule(lr{1pt}){2-4}
        & Expanded Prompt & 1.23&102.0 \\
        & Expanded Prompt + Ours &  \textbf{1.15}& \textbf{98.7}\\
        \addlinespace[2pt]
        \toprule
        \multirow{4}{*}{\shortstack{\textbf{Flux Dev} \\ (multi-step)}} & Base Prompt & 1.26&99.6\\
        & Base Prompt + Ours &  \textbf{1.21}& \textbf{97.7} \\
        \cmidrule{2-4}
        & Expanded Prompt & 1.27&99.5 \\
        & Expanded Prompt + Ours &  \textbf{1.23}& \textbf{97.6} \\
        \bottomrule
    \end{tabular}
    \caption{\textbf{Ablating image quality.} For the same samples evaluated in~\autoref{tab:ct2i}, we also compute their image quality, measured via CLIP Maximum Mean Discrepancy (CMMD)~\cite{jayasumana2024rethinking} and Frechet Inception Distance (FID)~\cite{heusel2017gans,parmar2021cleanfid}.}
    \label{tab:fid}
\end{table*}

\section{Additional Results for Visual Prompting}
\noindent \textbf{Horizon Control Evaluation.} Here, we design an automatic evaluation for horizon control.
We fine-tune LoRAs for each horizon line, and estimate the horizon of the generated images.
For our inputs, we take the cross product of three prompts from~\autoref{fig:horizon} and five lines approximately equally spaced across the image height.
For each LoRA, we generate images with five unseen seeds.
We use Perspective Fields~\cite{jin2022PerspectiveFields} to compute per-pixel latitude and longitude maps,
then extract the horizon from the zero-latitude line. We compute the ``horizon distance'' as the
difference between the estimated and control horizons, normalized by the image height.
We also validate the quality of this automatic metric by manually annotating 100 images,
where we find the mean error is 0.0470.
We report the results in~\autoref{tab:vp}, where we find that our method outperforms prompt expansion by 0.0218 points.

\begin{table}[ht]
\centering
\scriptsize
\begin{tabular}{lccc}
\toprule
& No Control & Prompt Expansion & Ours \\
\midrule
Horizon Distance ($\downarrow$) & 0.2204 & 0.1923 & \textbf{0.1705} \\
\bottomrule
\end{tabular}
\caption{\textbf{Quantifying horizon control.} On 75 images (3 prompts$\times$5 controls$\times$5 seeds), 
we report the horizon distance (bounded between 0 and 1) between the generated image and control. 
}
\label{tab:vp}
\end{table}

\noindent \textbf{Image Generator Prompts.} In~\autoref{tab:vp_prompts} we display the prompts associated with each figure in Sec.~\ref{subsec:visual_prompting} of the main text.

\begin{table}[h!]
    \begin{tabular}{@{}llp{12cm}@{}}
        \toprule
        Task & Figure & Prompts \\
        \midrule
        Color Palette &  \autoref{fig:palette} & \textit{``Pop art style household objects''}, \textit{``A pixel art sunset''}, \textit{``A colorful living room.''} \\
        Line Weight & \autoref{fig:line_weight} &  \textit{``Cartoon of a cat against a plain background.''} \\
        Horizon Position & \autoref{fig:horizon} &  \textit{``A teddy bear riding a skateboard in Times Square''}, \textit{``A dutch landscape painting''}, \newline\textit{``A tropical beach''} \\
        Relative Depth & \autoref{fig:depth} &  \textit{``A dog and a sheep.''} \\
        Visual Composition  & \autoref{fig:mondrian} &  \textit{``A painting of an oceanscape with a lighthouse, cruise, and seal colony''}, \newline\textit{``A painting of a Dutch landscape with a tree, windmill, and wheat field''} \\
        \bottomrule
    \end{tabular}
\phantomcaption
\label{tab:vp_prompts}
\end{table}

\begin{figure*}[!htb]
  \centering
  \includegraphics[width=\linewidth]{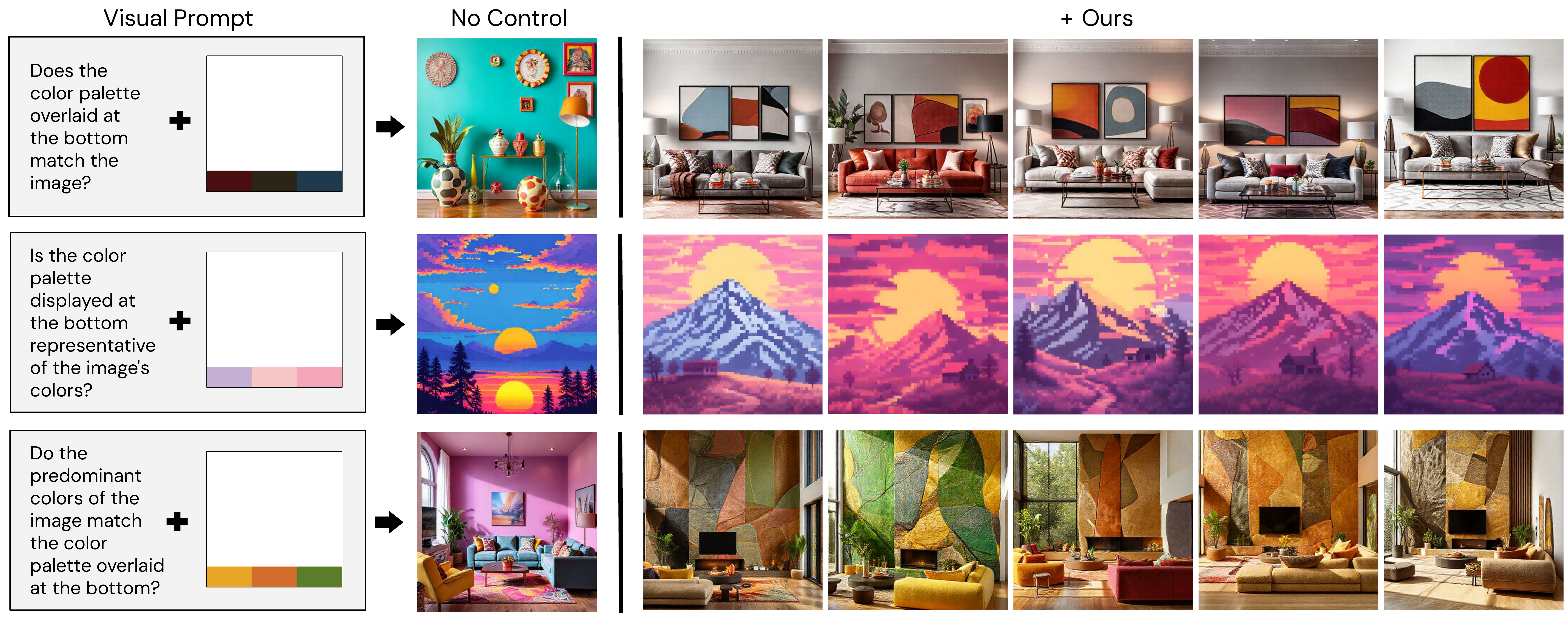}
  \captionof{figure}{\textbf{Color Palette.} Corresponding to~\autoref{fig:palette} in the main text, we show expanded results for a single set of weights optimized with our method, run on five random seeds.}
  \label{fig:full_palette}
  
  \vspace{1cm}

  \includegraphics[width=\linewidth]{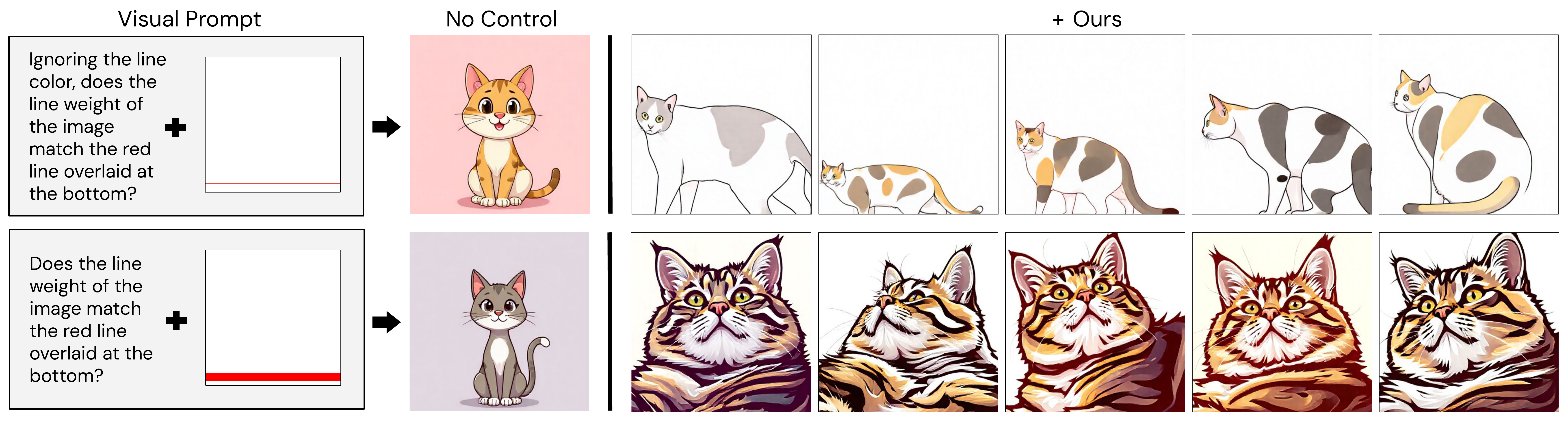}
  \captionof{figure}{\textbf{Line Weight.} Corresponding to~\autoref{fig:line_weight} in the main text, we show expanded results for a single set of weights optimized with our method, run on five random seeds.}
  \label{fig:full_line_weight}
\end{figure*}

\begin{figure*}
  \centering
  \includegraphics[width=\linewidth]{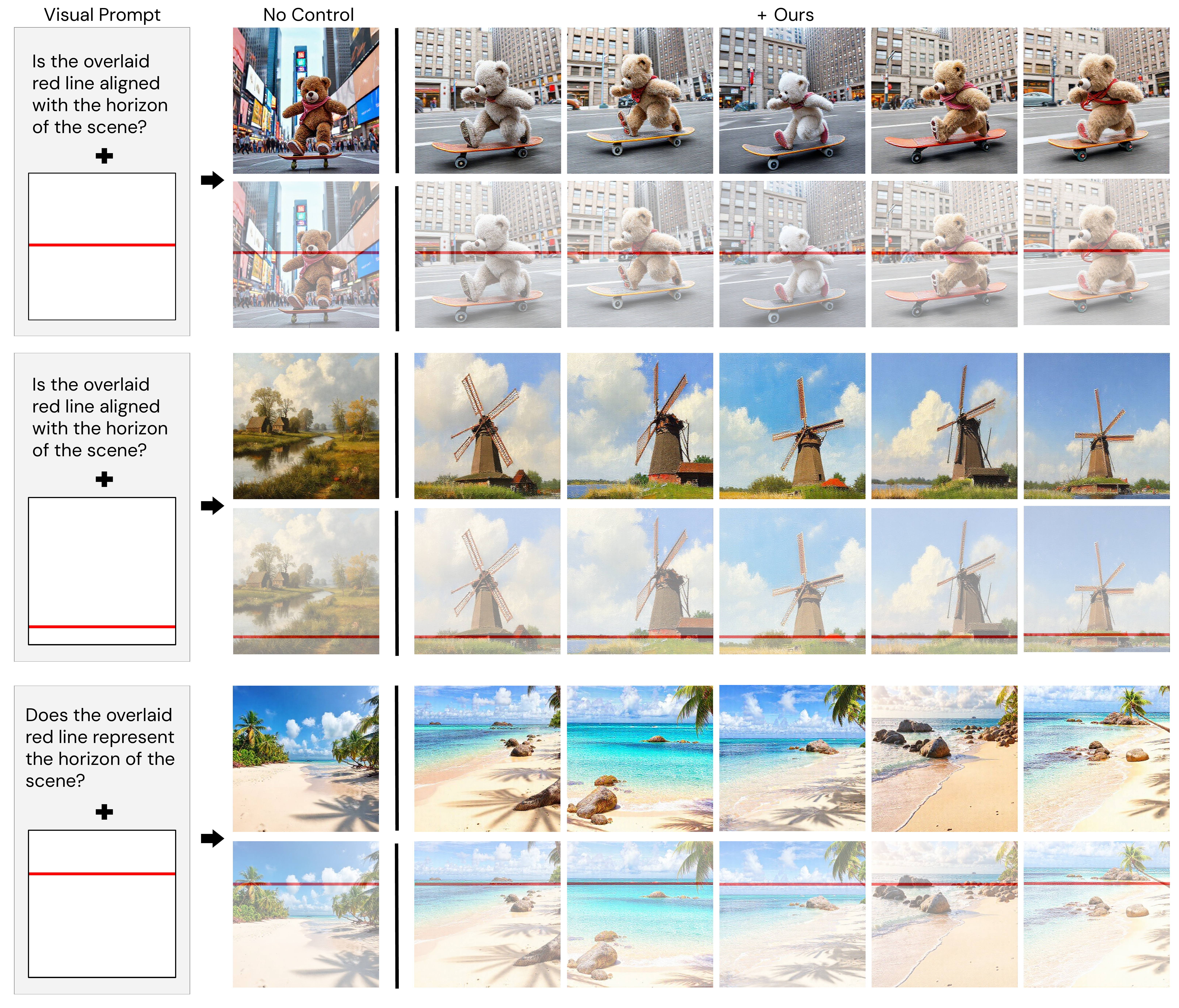}
  \caption{\textbf{Horizon Position.} Corresponding to~\autoref{fig:horizon} in the main text, we show expanded results for a single set of weights optimized with our method, run on five random seeds.}
  \label{fig:full_horizon}
\end{figure*}

\begin{figure*}
  \centering
  \includegraphics[width=\linewidth]{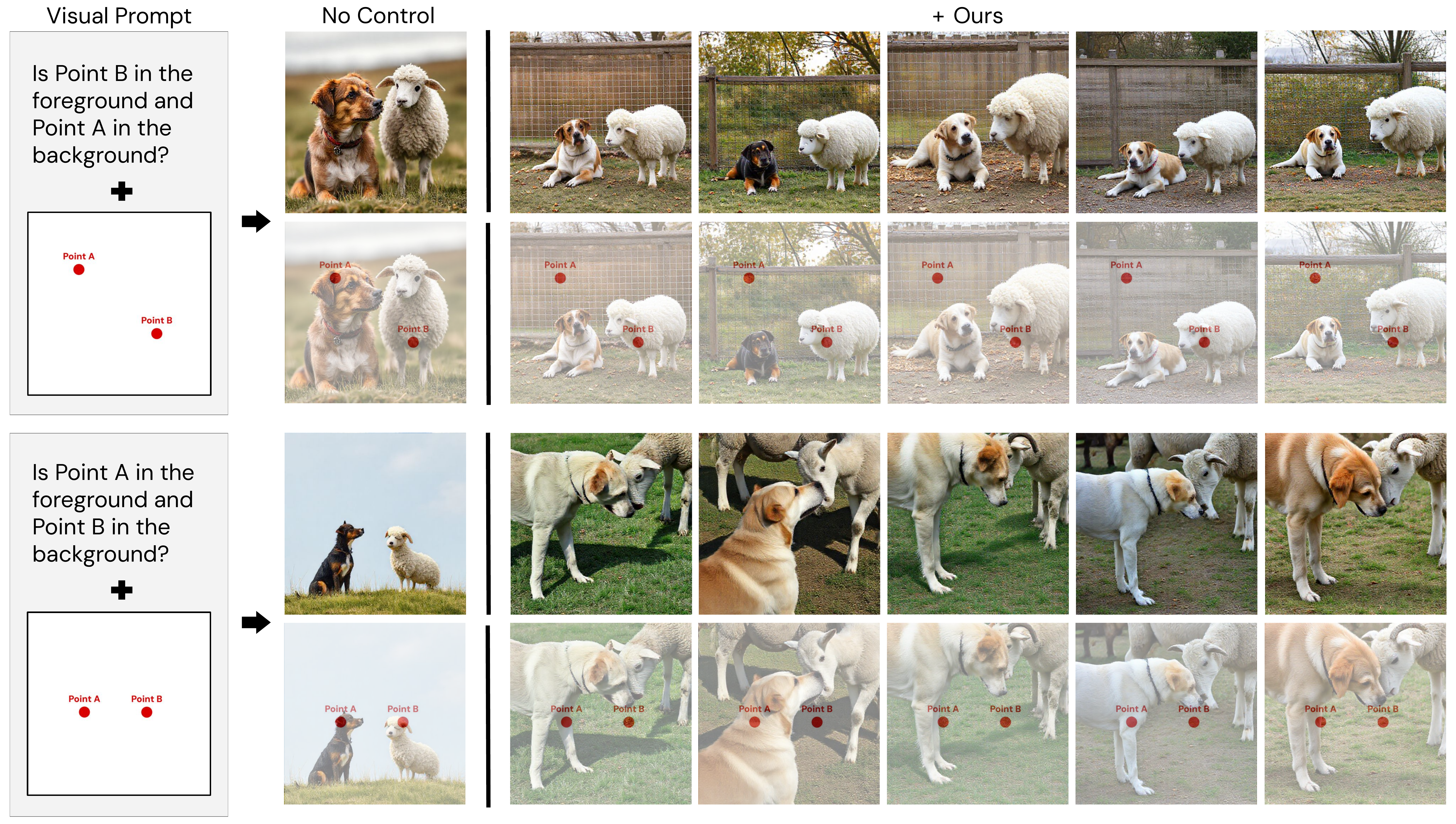}
  \caption{\textbf{Relative Depth.} Corresponding to~\autoref{fig:depth} in the main text, we show expanded results for a single set of weights optimized with our method, run on five random seeds.}
  \label{fig:full_depth}
\end{figure*}

\begin{figure*}
  \centering
  \includegraphics[width=\linewidth]{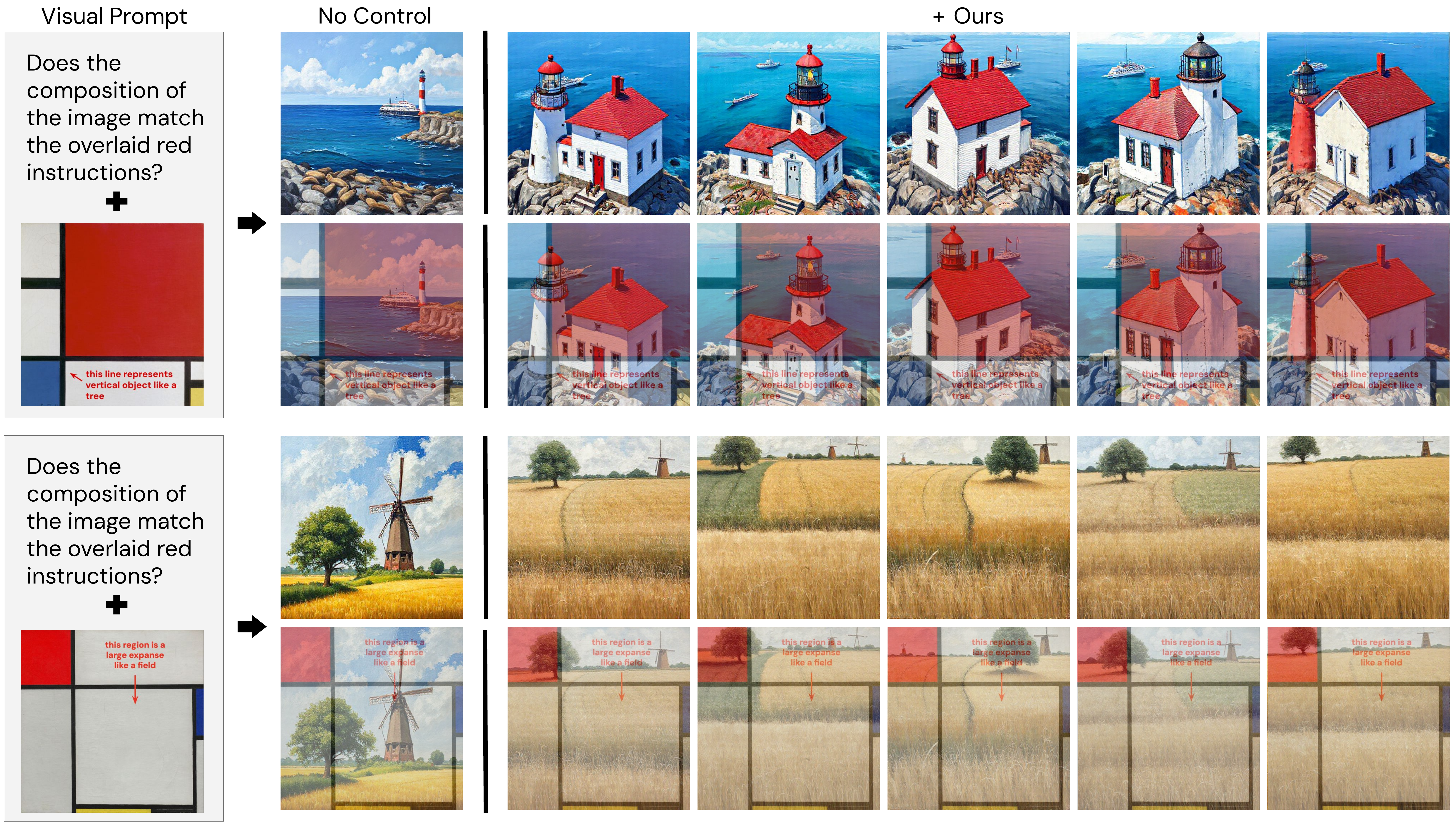}
  \caption{\textbf{Visual Composition.} Corresponding to~\autoref{fig:mondrian} in the main text, we show expanded results for a single set of weights optimized with our method, run on five random seeds.}
  \label{fig:full_mondrian}
\end{figure*}

\noindent \textbf{Additional Results.} In Figures~\ref{fig:full_palette}-\ref{fig:full_mondrian} we display extended results from our learned weights, run on five random seeds.

\newpage
\section{Templates and System Prompts for VLMs}
\textbf{Template Formatting.} As discussed in Sec.~\ref{sec:approach}, input formatting is important for obtaining meaningful scores from the VLM.
In Figures~\ref{fig:idefics-template},~\ref{fig:qwen-template} we display example inputs for different open-source VLMs, following their officially recommended templates.

\definecolor{specialmaroon}{RGB}{180, 60, 60}
\definecolor{specialblue}{RGB}{0, 102, 204}
\definecolor{specialimage}{RGB}{153, 51, 255}

\begin{figure*}[h!]
    \centering
    \begin{tcolorbox}
    \textbf{User:} \textcolor{specialimage}{\texttt{<image>}} \textcolor{specialblue}{Is the butter melting in the pan?} Answer with Yes or No.\texttt{<end\_of\_utterance>}\\
    \textbf{Assistant:} \textcolor{specialmaroon}{Yes}
    \end{tcolorbox}
    \caption{An example (\textcolor{specialimage}{image}, \textcolor{specialblue}{question}, \textcolor{specialmaroon}{answer}) triplet formatted for Idefics2~\cite{laurencon2024what}.}
    \label{fig:idefics-template}
\end{figure*}

\begin{figure*}[h!]
    \centering
    \begin{tcolorbox}
    \texttt{<|im\_start|>}\textbf{system} \\
    You are a helpful assistant.\\
    \texttt{<|im\_end|>} \\
    \texttt{<|im\_start|>}\textbf{user} \\
    \textcolor{specialimage}{\texttt{<|vision\_start|><|image\_pad|><|vision\_end|>}} \\
    \textcolor{specialblue}{Is the egg yolk and white still in liquid form?} Answer with Yes or No. \\
    \texttt{<|im\_end|>} \\
    \texttt{<|im\_start|>}\textbf{assistant} \\
    \textcolor{specialmaroon}{Yes}
    \end{tcolorbox}
    \caption{An example (\textcolor{specialimage}{image}, \textcolor{specialblue}{question}, \textcolor{specialmaroon}{answer}) triplet formatted for Qwen2.5-VL~\cite{Qwen2.5-VL}.}
    \label{fig:qwen-template}
\end{figure*}

\noindent \textbf{Prompt Expansion and Automatic Questions.} Here, we include additional details related to our commonsense understanding evaluation Sec.~\ref{subsec:ct2i}. For the prompt expansion baseline, we use the same system prompt used for DALLE-3~\cite{BetkerImprovingIG}, from~\citet{liu2024dalleprompt}. For the automatically generated questions used by our method, we design a system prompt shown in~\autoref{fig:question-generation}.

\newpage

\begin{figure*}
\centering
\scriptsize
\begin{tcolorbox}
You are an expert image verification assistant. Your task is to write questions that evaluate whether a generated image accurately reflects a given text prompt.

\vspace{1em}

\textbf{\#\#\# Instructions}
\begin{enumerate}
    \item You will be provided a prompt, and you will generate a list of questions.
    \item Each prompt is underspecified, in order to test a commonsense inference. 
    \item Think about the most likely inference the prompt is trying to test. This inference is related to physical laws, human practices, biological laws, daily items, and animal behaviors.
    \item Write a specific label for this inference.
    \item Write a question that checks this inference.
    \item The expected answer for all questions should be ``Yes''.
    \item Generate up to N=1 label-question-answer triplets. Express your output in JSON format as a list of question-answer pairs\\
    \verb|[[label1, question1, answer1], [label2, question2, answer2], ...]|
\end{enumerate}

\vspace{1em}

\textbf{\#\#\# Examples}\\
Labels and questions should be simple.\\
They should mention a minimal number of visible objects.\\
They should also pay attention to wording, for example, if an action is about to happen or has already happened.

\vspace{1em}

\textcolor{specialblue}{Prompt:} A straw submerged in a glass of water\\
\textcolor{specialmaroon}{Output:}
\begin{verbatim}
["Objects appear bent due to refraction.", 
 "Does the straw look bent underwater?", 
 "Yes"]
\end{verbatim}

\vspace{1em}

\textcolor{specialblue}{Prompt:} A firefighter on holiday\\
\textcolor{specialmaroon}{Output:}
\begin{verbatim}
["People wear casual clothes when they are off-duty.", 
 "Is the person wearing casual clothes?", 
 "Yes"]
\end{verbatim}

\vspace{1em}

\textcolor{specialblue}{Prompt:} An orange tree in the springtime\\
\textcolor{specialmaroon}{Output:}
\begin{verbatim}
["Orange trees have white blossoms when blooming.", 
 "Does the tree have white blossoms?", 
 "Yes"]
\end{verbatim}

\vspace{1em}

\textcolor{specialblue}{Prompt:} A balloon filled with air, tied to a fence\\
\textcolor{specialmaroon}{Output:}
\begin{verbatim}
["The balloon sinks because its air is the same density as the outside air.", 
 "Is the balloon on the ground?", 
 "Yes"]
\end{verbatim}

\vspace{1em}

\textcolor{specialblue}{Prompt:} A bird taking a nap\\
\textcolor{specialmaroon}{Output:}
\begin{verbatim}
["Birds sleep perched on branches in their natural habitat.", 
 "Is the bird asleep on a branch?", 
 "Yes"]
\end{verbatim}

\vspace{1em}

\textbf{\#\#\# Your Turn}\\
\textcolor{specialblue}{Prompt:} \texttt{<prompt>}\\
\textcolor{specialmaroon}{Output:}
\end{tcolorbox}
\caption{Here we display our system prompt used to automatically generate questions, to check commonsense inferences. To produce higher quality inferences, we include manually created in-context examples of \textcolor{specialblue}{Prompt}-\textcolor{specialmaroon}{Output} pairs separate from, but inspired by, the CommonsenseT2I~\cite{fu2024commonsenseti} benchmark.}
\label{fig:question-generation}
\end{figure*}

\end{document}